\documentclass[10pt,journal,compsoc]{IEEEtran}
\ifCLASSOPTIONcompsoc
  \usepackage[nocompress]{cite}
\else
  \usepackage{cite}
\fi
\ifCLASSINFOpdf
\else
\fi
\usepackage{amsmath}
\usepackage[pagebackref=true,breaklinks=true,letterpaper=true,colorlinks,bookmarks=false]{hyperref}

\usepackage{graphicx}
\usepackage{amssymb}
\usepackage{booktabs,caption} 
\usepackage{color}

\newcommand{\etal}{\textit{et al}. }

\begin{document}
%
\title{FaceScape: 3D Facial Dataset and Benchmark for Single-View 3D Face Reconstruction}
%
%
%
%

\author{Hao~Zhu$^{*}$,~\IEEEmembership{Member,~IEEE}, Haotian~Yang$^{*}$, 
        Longwei~Guo, Yidi~Zhang, Yanru~Wang, Mingkai~Huang, Menghua~Wu,
        Qiu~Shen, Ruigang~Yang,~\IEEEmembership{Fellow,~IEEE} 
        and~Xun~Cao,~\IEEEmembership{Member,~IEEE}

\IEEEcompsocitemizethanks{
\IEEEcompsocthanksitem This work was supported in part by the National Key Research and Development Program of China under Grant 2022YFF0902401, in part by the National Natural Science Foundation of China under Grants 62001213, 62025108, 62071216, 62231002, and U1936202, in part byTencentRhino-Bird Research Program, and in part by gift funding from Huawei. Recommended for acceptance by K. M. Lee (EIC). (Hao Zhu and Haotian Yang contributed equally to this work.) (Corresponding author: Xun Cao.)
\IEEEcompsocthanksitem Hao Zhu, Haotian Yang, Longwei Guo, Yidi Zhang, Yanru Wang, Mingkai Huang, MenghuaWu, Qiu Shen, and Xun Cao are with the Nanjing University, Nanjing 210093, China (e-mail: zh@nju.edu.cn; yanght321@gmail.com; guolongwei@smail.nju.edu.cn; zyd@smail.nju.edu.cn; wyr@smail.nju.edu.cn; mg1723073@smail.nju.edu.cn; menghuawu@smail.nju.edu.cn; shenqiu@nju.edu.cn; xuncao@nju.edu.cn).
\IEEEcompsocthanksitem RuigangYang is with the University of Kentucky, Lexington, KY40506 USA (e-mail: ryang2@uky.edu)
\IEEEcompsocthanksitem Code \& data are available at \url{https://github.com/zhuhao-nju/facescape.git}.

}
}

\IEEEtitleabstractindextext{%
\begin{abstract}
In this paper, we present a large-scale detailed 3D face dataset, \emph{FaceScape}, and the corresponding benchmark to evaluate single-view facial 3D reconstruction. By training on FaceScape data, a novel algorithm is proposed to predict elaborate riggable 3D face models from a single image input.
FaceScape dataset releases $16,940$ textured 3D faces, captured from $847$ subjects and each with $20$ specific expressions. The 3D models contain the pore-level facial geometry that is also processed to be topologically uniform. These fine 3D facial models can be represented as a 3D morphable model for coarse shapes and displacement maps for detailed geometry.
Taking advantage of the large-scale and high-accuracy dataset, a novel algorithm is further proposed to learn the expression-specific dynamic details using a deep neural network. The learned relationship serves as the foundation of our 3D face prediction system from a single image input. Different from most previous methods, our predicted 3D models are riggable with highly detailed geometry under different expressions. 
We also use FaceScape data to generate the in-the-wild and in-the-lab benchmark to evaluate recent methods of single-view face reconstruction. The accuracy is reported and analyzed on the dimensions of camera pose and focal length, which provides a faithful and comprehensive evaluation and reveals new challenges.
The unprecedented dataset, benchmark, and code have been released to the public for research purpose$^{1}$.

\end{abstract}

}

\maketitle

\begin{figure*}[t]
\begin{center}
    \includegraphics[width=1.0\linewidth]{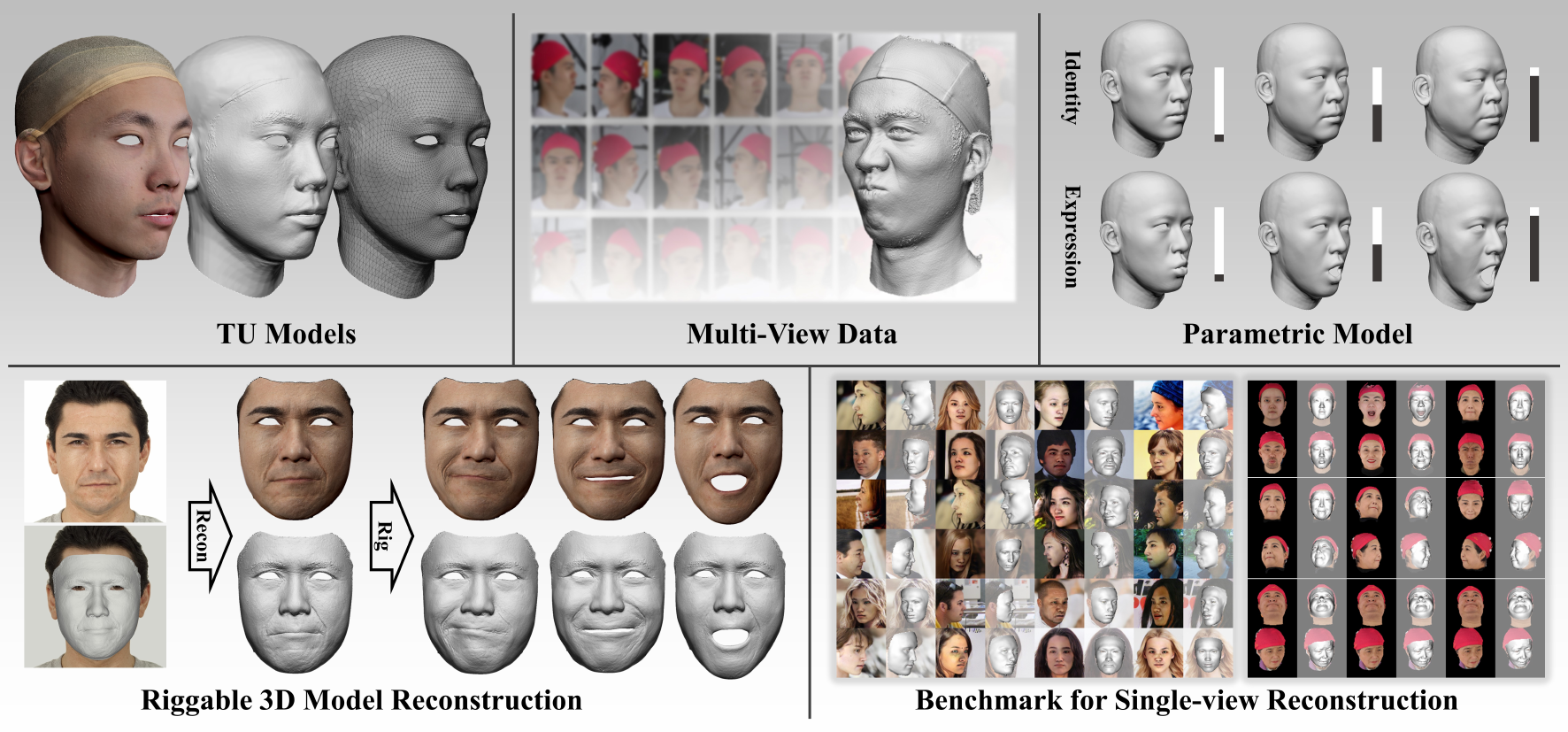}
\end{center}
    \vspace{-0.2in}
    \caption{FaceScape contains a large-scale detailed 3D face dataset and the benchmarks to evaluate single-view facial 3D reconstruction. By training on FaceScape data, a novel algorithm is proposed to predict elaborate riggable 3D face models from a single image input.}
\label{fig:teaser}
\end{figure*}

\IEEEdisplaynontitleabstractindextext

%
\IEEEpeerreviewmaketitle


%
%
%
%

 

\section{Introduction}

\IEEEPARstart{P}{arsing} and recovering 3D face models from images have been hot research topics in both computer vision and computer graphics due to their many applications. As learning-based methods have become mainstream in face tracking, recognition, reconstruction, and synthesis, 3D face datasets become increasingly important.  Though numerous 2D face datasets are released, there are few datasets containing large-scale and high-quality 3D faces. As such, the researches of facial parsing and reconstruction that rely on 3D faces suffer. 

Existing 3D face datasets capture the face geometry using sparse camera array\cite{booth20163d, dai20173d, phillips2005overview} or active depth sensors such as Kinect\cite{cao2013facewarehouse} and coded light\cite{paysan20093d}.  These setups limit the quality of the recovered faces.  We captured the 3D face model using a dense 68-camera array, which recovers the 3D face model with wrinkles and pore-level detailed geometry, as shown in Figure~\ref{fig:teaser}.  In addition to shape quality, our dataset provides a considerable amount of scans for study.  We invited $897$ participants between the ages of $17$ and $80$ as subjects, of which $847$ subjects are released for generating parametric models and training prediction models. Each subject is guided to perform $20$ specified expressions, generating $17,940$ high-quality 3D face models. The corresponding color images and subjects' information (such as age and gender) are also recorded.

Based on the high-fidelity 3D raw model, we build a powerful parametric model to represent the detailed face shape.  All the raw scans are first transformed into a topologically uniformed model (TU model) representing the coarse shape and a displacement map representing detailed shapes. The TU models are further used to build bilinear models in identity and expression dimensions.  Experiments show that our generated bilinear model outperforms previous methods in terms of representation capability.

Using the FaceScape dataset, we study how to predict a detailed riggable face model from a single image.  
The main problem is how to predict the variation of small-scale geometry caused by facial expression movement, such as wrinkles.  We propose the dynamic details which can be predicted from a single image and can be learned from the FaceScape dataset. Cooperated with the bilinear model fitting algorithm, a two-stage pipeline to predict a detailed riggable face is presented.  Firstly, a coarse mesh model is fitted according to the detected 2D landmarks; Secondly, displacement maps for each expression are predicted, which constitutes a linear space representing mesoscopic geometry in any expression. Different from prior works~\cite{tran2018extreme, chen2019photo, chen2020self} that predict static geometric details, our predicted details can be rigged into any expressions. DECA~\cite{feng2021learning} can predict animatable shape details and learn the person-specific and expressions-dependent details from in-the-wild images. By contrast, our method represents dynamic details with a linear combination of a set of displacement maps, similar to blendshapes, then learn the nonlinear mapping from a source image to this linear combination, with our FaceScape datasets. The experiments show that our predicted 3D face can be rigged to various expressions containing unseen wrinkles and other details. 

A new benchmark containing in-the-wild and in-the-lab data is presented to evaluate single-view face reconstruction. 14 recent methods are evaluated on the dimensions of camera pose and focal length, which provides a comprehensive evaluation and reveals new challenges.

Our contributions are shown in Figure~\ref{fig:teaser}, and are summarized as following:
\begin{itemize} 
	\item We release a large-scale 3D face dataset, FaceScape, consisting of $16,940$ extremely detailed 3D face models. The raw scans are processed into TU models representing coarse geometry and displacement maps representing detailed geometry. The dataset has been publicly released for non-commercial research purposes.
 
	\item A two-stage pipeline is presented to recover a detailed riggable face model from a single image. We model the detailed variation of geometry across facial movement as dynamic details, which can be learned from the FaceScape dataset. The predicted 3D face can be rigged to various expressions containing unseen wrinkles and other details.
	
	\item A new benchmark containing in-the-wild and in-the-lab data is presented to evaluate single-view face reconstruction (SVFR) methods, which reviews the SOTA methods on multiple dimensions and reveals new challenges.
	
\end{itemize} 

\section{Related Work}
\label{sec:related}

\subsection{3D Face Acquisitions and Datasets}

3D face acquisition has long been a subject of great concern while modeling accurate 3D faces is non-trivial. 3D face acquisition methods can be generally categorized into active methods that leverage an optical signal transmitter and passive methods that merely leverage common cameras. 
Early attempts~\cite{savran2008bosphorus, baocai2009bjut, sankowski2015multimodal} leverage a 3D laser scanner to obtain the detailed 3D facial shape, while the main drawback is the long scanning duration that may bring error caused by the minor shake of the subjects. 
In later researches\cite{yin20063d, cao2013facewarehouse, cheng20184dfab, wang2022faceverse, dai2020statistical, bagdanov2011florence}, structured light scanners, stereo cameras, and depth cameras overcome the disadvantages of speed, but the spatial resolution of these acquisition devices is limited. 
In short, the aforementioned active 3D reconstruction methods obtain accurate facial depth at the expense of running speed or spatial resolution.

On the other hand, increasing efforts are made to capture 3D faces with passive methods, where high spatial resolution and real-time running are achievable with sophisticated algorithms. Sparse multi-view camera arrays are adopted to build several 3D face datasets~\cite{lattas2020avatarme, lattas2021avatarme++, zhang2013high, cosker2011facs}, however, they suffer from unstable performance since limited correspondences can be recovered from the sparse-view images for 3D reconstruction. 
Except for the multi-view camera arrays that have been adopted for creating 3D face datasets, cutting-edge passive algorithms for facial geometry recovery are still updated rapidly, of which the accuracy and detail-fidelity are obviously beyond that of 3D scanners and depth cameras~\cite{beeler2010high, ghosh2011multiview, fyffe2016near, ghosh2011multiview, stratou2011effect, alexander2010digital, xiao2022detailed}. These methods have been applied to movie production and high-fidelity digital human creation, but due to the high equipment cost and the complexity of portrait licensing, there is no large-scale 3D face data set established by such methods before our work.

In addition to 3D reconstruction methods, some researchers create 3D face datasets by fitting a statistical 3D face model to in-the-wild facial images\cite{paysan20093d, zhu2016face, guo2018cnn, booth20173d, booth20183d}. This strategy makes it convenient to build a large-scale dataset as the 2D images containing faces are easy to be collected and was widely used in training a learning-based 3D face predictor. The major problem of the fitted 3D face datasets is that the fitted 3D faces may contain biased features caused by limited statistical data. Besides, 2D images and 3D shapes in the fitted datasets are distinctly misaligned. These uncertainties make the existing fitted datasets unable to completely replace the datasets created by 3D reconstruction but serve as an augmentation.

In this work, we captured 3D faces using passive reconstruction methods with a multi-view camera array. Different from the previous methods, we use a dense camera array that contains $68$ DSLR cameras. This setup enables pore-level geometry capture in an instant shot. Our dataset outperforms previous publicly available datasets in terms of quality and quantity: pore-level geometry/texture and $897$ identities $\times$ $20$ expressions  $= 17,940$ static 3D models.

\subsection{Parametric 3D Face.}

A parametric 3D face is a statistical model which transforms the shape and texture of a number of 3D faces into a parametric space, and a classic one is known as 3D morphable model (3DMM)~\cite{blanz1999morphable}.  
The representation capability and feature disentangling ability are the two most important indicators of parametric 3D faces. The former determines whether a parametric face can represent more diverse and detailed faces, and the latter determines the application range of a parametric face in parsing, editing, and rigging. 

In terms of representation capability, early parametric models leverage a linear mapping model which can be created by principle content analysis (PCA) methods~\cite{blanz1999morphable, vlasic2005face, patel20093d, cao2013facewarehouse}, while subsequent studies have shown that a nonlinear model combined with prior knowledge can significantly improve the representation capability~\cite{tewari2017mofa, bagautdinov2018modeling, tewari2018self, tran2019towards, tran2019learning, tran2018nonlinear, abrevaya2018multilinear, ranjan2018generating, lombardi2018deep, cheng2019meshgan, bouritsas2019neural, yenamandra2021i3dmm, wu2023high, zhuang2022mofanerf}. Among them, deep neural networks are representative nonlinear models and have shown superior performance in representing detailed 3D faces. In terms of feature disentangling ability, recent works explore disentangling the parametric space into multiple dimensions including identity, expression, and visemes, so that the parametric model can be controlled by these attributes separately\cite{vlasic2005face, cao2013facewarehouse, brunton2014multilinear, bolkart20153d, bolkart2016robust, li2017learning, abrevaya2018multilinear, jiang2019disentangled, cudeiro2019capture}.  The models in the expression dimension could be further transformed to a set of blendshapes~\cite{li2010example}, which can be rigged to produce individual-specific animation.  
We recommend referring to the recent survey\cite{egger20203d} for a comprehensive review of 3D morphable models.

Inspired by the prior works, we build a two-stage parametric 3D face that treats coarse shapes and fine shapes separately. The coarse shapes and colors are modeled by a linear model with identities and expressions decoupled. The fine shapes are transformed into UV displacement maps and are modeled by convolutional neural networks. 

\subsection{Single-View Face Reconstruction}

In recent years, single-view face reconstruction (SVFR) has attracted a lot of attention and developed rapidly.
Generally, SVFR methods can be categorized into parametric methods and non-parametric methods.  Parametric methods treat the reconstruction of facial shape as a parametric model fitting problem, which optimizes or predicts a vector representing the 3D face from a single image\cite{romdhani2005estimating, amberg2007optimal, thies2016face2face, zhu2016face, fried2016perspective, dou2017end, tewari2017mofa, deng2019accurate, shang2020self, guo2020towards, sanyal2019learning, tran2018nonlinear, tran2019towards, tewari2018self, genova2018unsupervised, koizumi2020look, liu2018disentangling, gecer2019ganfit, tu20203d}. These methods are also known as 3D face alignment. Considering that the parametric model cannot recover detailed shape due to the limited representation capability, the latter methods propose to first reconstruct a coarse 3D face with parametric methods, then predict detailed geometry based on the coarse shape\cite{richardson2017learning, sela2017unrestricted, sengupta2018sfsnet, huynh2018mesoscopic, chen2020self, chen2019photo, tran2018extreme, feng2021learning}.  
Non-parametric methods directly predict the facial shape in the form of mesh vertices~\cite{feng2018joint, ruan2021sadrnet, zeng2019df2net, sela2017unrestricted, zhou2019dense, zhu2020beyond}, depth~\cite{zhang2021learning}, 3D volume~\cite{jackson2017large}, or implicit functions~\cite{yenamandra2021i3dmm, zhuang2022mofanerf, guo2023rafare}. 
Compared to parametric models, non-parametric representation owns higher degrees of freedom and therefore can express finer shape details. However, the free-form representation brings less constraint and makes it difficult to predict the accurate shape. We recommend referring to the survey\cite{knothe2011morphable} for a comprehensive review of fitting 3DMM to a single image.

In this paper, we propose to recover the coarse base shape using parametric methods and predict detailed and riggable details using non-parametric methods. We demonstrate that a detailed and rigged 3D face model can be recovered from a single image. The rigged model exhibits expression-depended geometric details such as dynamic wrinkles.

\section{Dataset}

\subsection{3D Face Capture}

\begin{figure}[t]
\begin{center}
    \includegraphics[width=1.0\linewidth]{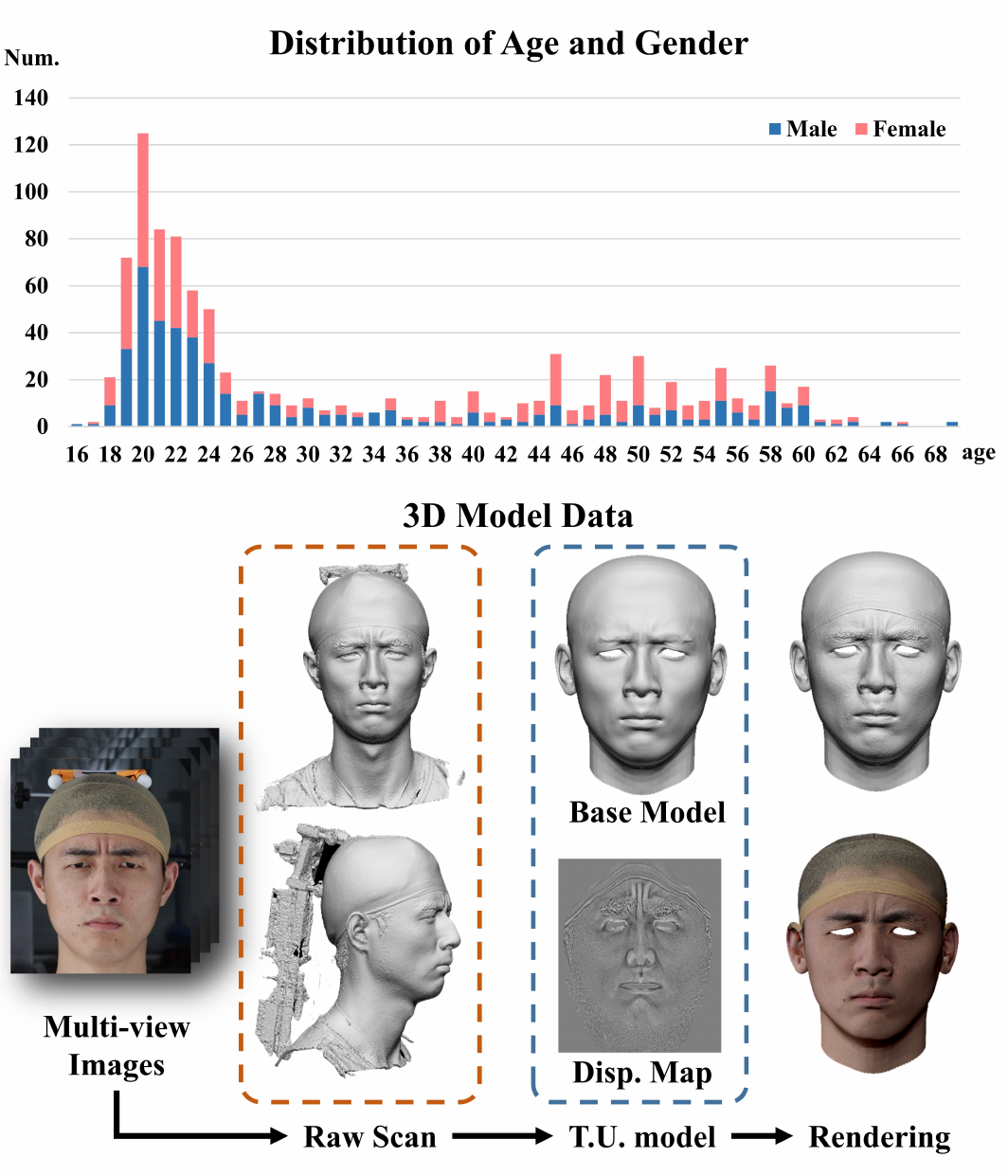}
\end{center}
    \vspace{-0.25in}
    \caption{Description of FaceScape dataset.  On the upper side, we show the histogram of the subjects' age and gender.  On the lower side, we show the pipeline from the captured multi-view images to topologically uniformed models. }
\label{fig:statistic}
\end{figure}

We use a multi-view 3D reconstruction system to capture the raw mesh model for the datasets.  The multi-view system consists of 68 DSLR cameras, 30 of which capture 8K images focusing on the front side, and the other cameras capture 4K level images for the side part.  The camera shutters are synced to be triggered within $5$ms.  We spend six months inviting 897 people to be our capturing subjects.  The subjects are between 16 and 70 years old and are mostly from Asia.  We follow FaceWarehouse\cite{cao2013facewarehouse} which asks each subject to perform 20 specific expressions including the neutral expression for capturing.  The total reconstructed number reaches roughly $17,940$, which is the largest amount compared to previous expression-controlled 3D face datasets.  The reconstructed model is a triangle mesh with roughly 2 million vertices and 4 million triangle faces.
The meta-information for each subject is recorded voluntarily, including age, gender, and job.  We show the statistical information about the subjects in our dataset in Figure~\ref{fig:statistic}.

\subsection{3D shape Registration}

We down-sample the raw recovered mesh into coarse mesh with fewer triangle faces, namely base shape, and then build 3DMM for these simplified meshes.  First, we register all the meshes to the template face model roughly by aligning 3D facial landmarks, then the NICP\cite{amberg2007optimal} is used to deform the templates to fit the scanned meshes.  The deformed meshes can be used to represent the original scanned face with minor accuracy loss, and more importantly, all of the deformed models share a uniform topology. The detailed steps to register all the raw meshes are described in the supplementary material.

After obtaining the topology-uniformed base shape, we use displacement maps in UV space to represent middle and fine-scale details that are not captured by the base model due to the small number of vertices and faces. 
We find the surface points of the base mesh corresponding to the pixels in the displacement map, then inverse-project the points to the raw mesh along normal directions to find its corresponding points. The pixel values of the displacement map are set to the signed distance from the point on the base mesh to its corresponding point.  

We use base shapes to represent coarse geometry and displacement maps to represent detailed geometry, which is a two-layer representation of our extremely detailed face shape.  The new representation takes roughly $2\%$ of the original mesh data size while maintaining the mean absolute error to be less than $0.3$mm.

\subsection{Appearance-based Refinement}

The 3D registration mentioned above only utilizes the geometry of the topologically uniformed models to align mesh models. Though the registered shape is aligned with the raw scan, it fails to recover stretching deformations such as eyebrows raising and mouth stretching. Besides, the registered meshes may deviate from the semantic position of the texture in the flexible area like the mouth. The reason is that NICP only tries to minimize the nearest distance, disregarding the motions that are parallel to the surface. To solve these problems, we use the optical flow in UV texture space to refine the registration results. 

In previous studies, landmarks-based warping using thin plate splines (TPS) algorithm~\cite{bookstein1989principal} is a more commonly-used way to align 3D faces and build parametric face models~\cite{patel20093d}.   However, we find that flow-based refinement is a better solution for our task. The major reason is that our refine aims compute the appearance correspondence of the same person, and with the uniform illumination in our capture system, it is quite robust to retrieve dense and accurate correspondences with optical flow. On the contrary, the prior works aim to align the faces of different people, whose appearance may differ a lot. So landmarks-based refinement was leveraged to find a stable but sparse correspondence.

\begin{figure}[t]
\begin{center}
    \includegraphics[width=1.0\linewidth]{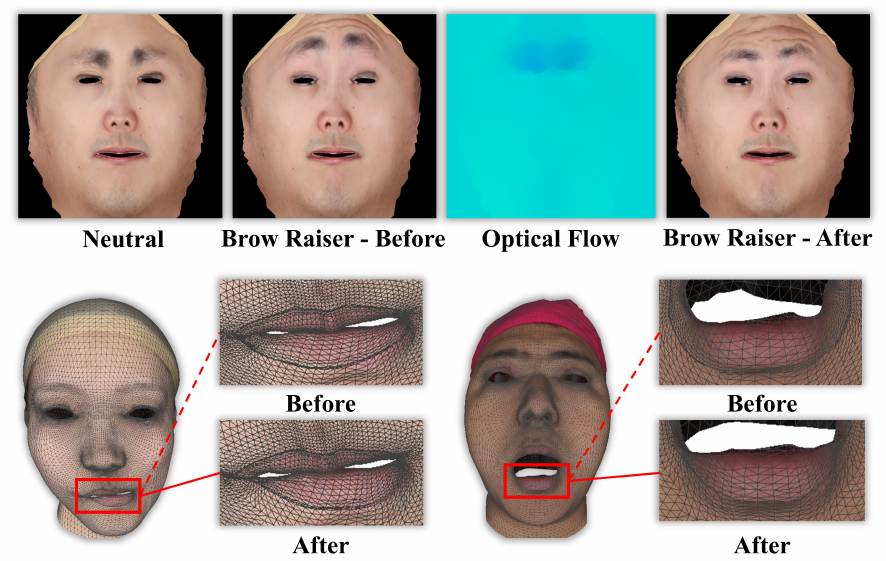}
\end{center}
    \caption{In the first row, the texture shift between `Neutral' and `Brow Raiser' can be estimated as the optical flow, then the mesh grid can be refined with the updated texture shown on the right. In the second row, we compare the result before appearance-based refinement and after appearance-based refinement. The grid of lips is misaligned in `Before', and is fixed in `After'.}
\label{fig:reg_refine}
\end{figure}

We observe that the textures of a certain subject should be consistent in different expressions. However, due to the misalignment in the registration process, the texture maps of different expressions will shift, as shown in Figure~\ref{fig:reg_refine}. We project and blend the pixels of the multi-view images based on the preliminary registration to obtain the texture. Then the optical flow is estimated with DeepFlow~\cite{weinzaepfel2013deepflow} from the texture of other expressions to the neutral expression, and the vertex position of the 3D mesh is refined according to the optical flow.

Let $M_{n}$ be the preliminary registration in a neutral position and $M_{e}$ in another expression, and the corresponding textures are $I_{n}$ and $I_{e}$. The optical flow $F$ from $I_{n}$ to $I_{e}$ is computed by DeepFlow, which represents the dense correspondence between $I_0$ and $I_1$. The pixel with the coordinate $(x,y)$ in $I_n$ correspond to the pixel in $I_1$ with coordinate $(x-F(x,y,0),y-F(x,y,1))$. Based on the correspondence, we can obtain the updated vertices $\hat{V}_e'$ in $M_{e}$. That is, the vertices with texture coordinate $(x,y)$ should have the same position as the vertices on $M_{e}$ with texture coordinate $(x-F(x,y,0),y-F(x,y,1))$, so that the texture of $M_{e}$ is consistent to the neutral expression. The position corresponding to $(x-F(x,y,0),y-F(x,y,1))$ is obtained by barycentric interpolation of triangle vertices. Due to the defect of optical flow calculation, directly updating the vertices' position will lead to a twisted mesh. So we add a smooth regularization item to ensure adjacent vertices do not change dramatically. The energy function to optimize the new vertices $\hat{V}$ is formulated as:

\begin{equation}
\begin{aligned}
\hat{V}=\mathop{\arg\min}_{\hat{V}} ||\hat{V}-\hat{V}'||^2+\sum_{(i,j)\in \Omega}||(\hat{V}_i-{\hat{V}'}_i)-(\hat{V}_j-{\hat{V}'}_j)||^2
\end{aligned}
\end{equation}

where $\Omega$ are the edges in the mesh. 

We show the registration before and after appearance-based refinement in Figure~\ref{fig:reg_refine}. Though the shape of the preliminary registration is close to the raw scan, the semantic meaning of the triangles and texture is misaligned, which is significantly alleviated after the refinement. More experiments and discussions about appearance-based refinement are shown in the supplementary material.

\subsection{Bilinear Model}\label{sec:3dmm}

Bilinear model\cite{vlasic2005face} is a special form of the 3D morphable model to parameterize face models in both identity and expression dimensions. The bilinear model can be linked to a face-fitting algorithm to extract identity, and the fitted individual-specific model can be further transformed into riggable blendshapes. 
Here we describe how to generate the bilinear model from our topologically uniformed models.  
Given 20 registered meshes in different expressions, we use the example-based facial rigging algorithm\cite{li2010example} to generate 52 blendshapes based on FACS\cite{Ekman1978FacialAC} for each person. Then we follow the previous methods\cite{vlasic2005face, cao2013facewarehouse} to build the bilinear model from generated blendshapes in the space of 26317 vertices $\times$ 52 expressions $\times$ 847 identities. Specifically, we use Tucker decomposition to decompose the large rank-3 tensor to a small core tensor $C_r$ and two low-dimensional components for identity and expression. A refined face shape can be generated given the identity parameter $\textbf{w}_{id}$ and expression parameter $\textbf{w}_{exp}$ as:
\begin{equation}
\begin{aligned}
\mathop{V}=C_r \times \textbf{w}_{exp} \times \textbf{w}_{id}
\end{aligned}
\end{equation}
where $V$ is the vertex position of the generated mesh. 

The superiority in quality and quantity of the FaceScape dataset determines the larger representation capability of our bilinear model, which will then be verified in Section~\ref{sec:exp_represent}.

\section{Detailed Riggable Model Prediction}

\begin{figure}[t]
\begin{center}
    \includegraphics[width=1.0\linewidth]{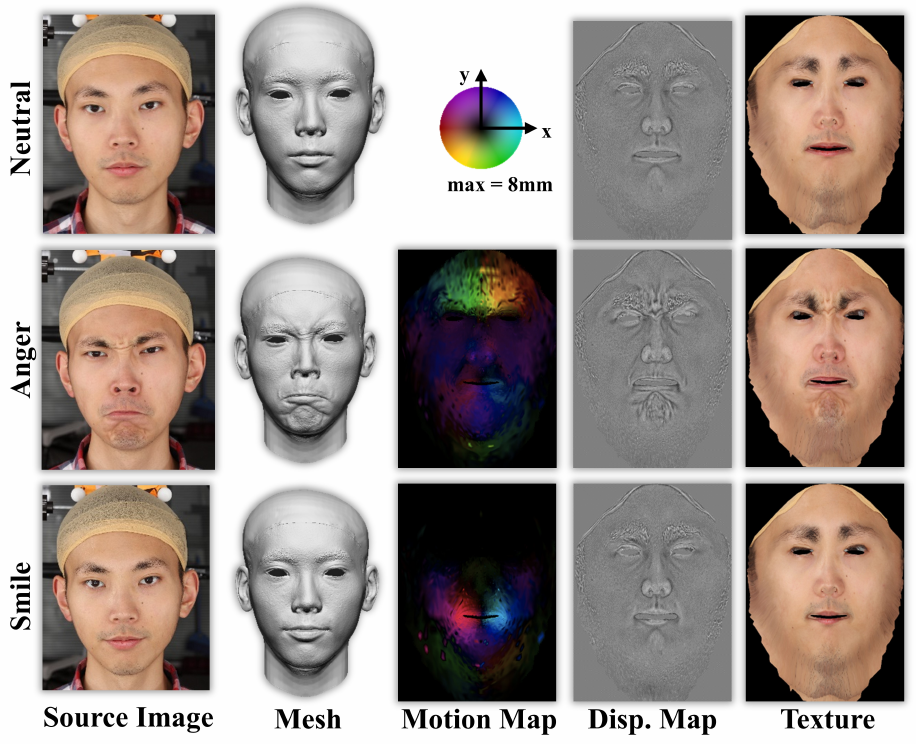}
\end{center}
    \vspace{-0.2in}
    \caption{Riggable details can be decoupled as static details and dynamic details.  The static details can be estimated from the facial textures, while the dynamic details are strongly related to the facial deforming map.
    }
\label{fig:detail}
\end{figure}

\begin{figure*}[t]
\begin{center}
    \includegraphics[width=1.0\linewidth]{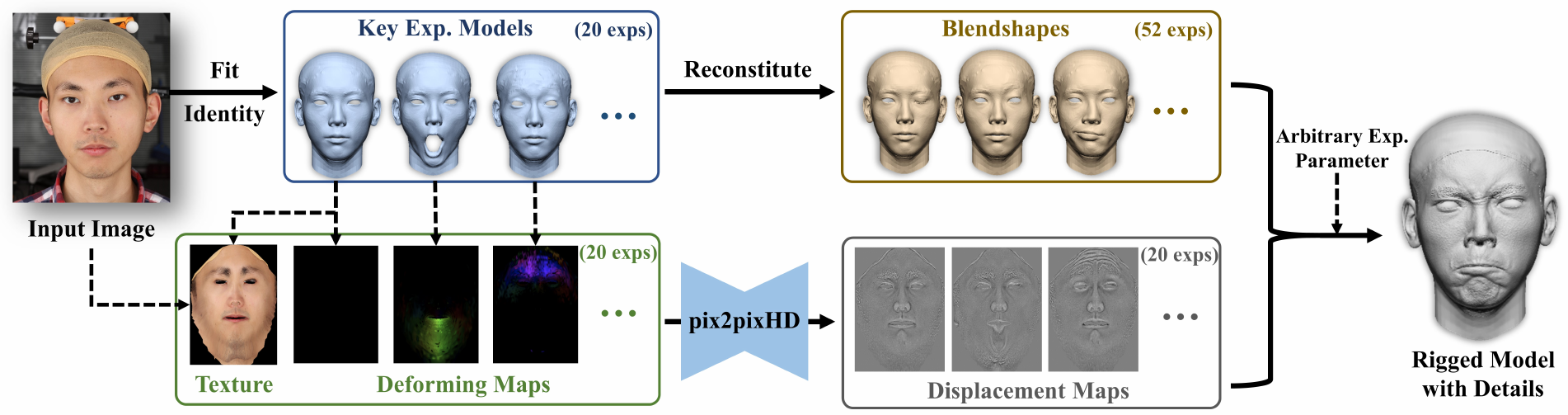}
    \vspace{-0.3in}
\end{center}
    \caption{The pipeline to predict a detailed riggable 3D face from a single image consists of base model fitting(blue box), displacement map prediction (green and grey box), and dynamic details synthesis (yellow box and the followed arrow on the right). 
    }
\label{fig:method}
\end{figure*}

As reviewed in the related works in Section~\ref{sec:related}, existing methods have succeeded in recovering a detailed 3D facial model from a single image, while the results of most previous methods are not riggable with highly detailed geometry under different expressions.  
The FaceScape dataset makes it possible to estimate detailed and riggable 3D face models from a single image, as we can learn the dynamic details from a large amount of detailed facial models.  We show our pipeline in Figure~\ref{fig:method} to predict a detailed and riggable 3D face model from a single image.  The pipeline consists of three stages: base model fitting, displacement map prediction, and dynamic details synthesis.  We will explain each stage in detail in the following sections.

\subsection{Base Model Fitting}
\label{sec:bmfit}

The bilinear model for base shape is inherently riggable as the parametric space is separated into the identity dimension and expression dimension, so the coarse riggable model can be generated by regressing the parameters of identity for the bilinear model. Following \cite{thies2016face2face}, we estimate parameters corresponding to a given image by optimizing an objective function consisting of three parts.
The first part is the landmark alignment term. Under the assumption of a weak perspective camera model, the landmark alignment term $E_{lan}$ is defined as the distance between the detected 2D landmark and its corresponding vertex projected on the image space. The second part is pixel-level consistency term $E_{pixel}$ measuring how well the input image is explained by a synthesized image. The last part is a regularization term that formulates identity, expression, and albedo parameters as multivariate Gaussians.
The final objective function is given by:
\begin{equation}
\begin{aligned}
E=E_{lan}+\lambda_{1} E_{pixel}+{\lambda_2 E_{id}}+{\lambda_3 E_{exp}}+{\lambda_4 E_{alb}}
\end{aligned}
\end{equation}
where $E_{id}$, $E_{exp}$ and $E_{alb}$ are the regularization terms of expression, identity and albedo, respectively. $\lambda_1$, $\lambda_2$, $\lambda_3$ and $\lambda_4$ are the weights of different terms. 

After obtaining the identity parameter $w_{id}$, individual-specific blendshapes $B_i$ can be generated as:
\begin{equation}
\begin{aligned}
B_i=C_r \times \hat{w_{exp}}^{(i)} \times w_{id}, 0\leq i\leq 51
\end{aligned}
\end{equation}
where $\hat{w_{exp}}^{(i)}$ is the expression parameter corresponding to blendshape $B_i$ from Tucker decomposition. 

\textbf{Learning-based fit.} As an alternative to the optimization-based fit as explained above, we can train a neural network to regress the parameters of the bilinear model and camera pose.  We use ShuffleNet\_v2\cite{ma2018shufflenet} as our regressor, which takes a $256\times256$ image as input and predicts a $107$-dimension vector.  The predicted vector consists of 50 identity parameters, 51 expression parameters, rotation quaternion with scale (4 parameters), and translation in image coordinate (2 parameters). We use $17,810$ images from CelebA\cite{liu2015faceattributes} as a training set and generate the parameters label using the optimization-based fitting as explained above. Learning-based fit is much faster than the optimization-based method and achieves comparable accuracy. However, the learning-based fit is unstable for some specific poses such as head raising and bowing, because these poses are rare in the training set generated from CelebA\cite{liu2015faceattributes}.  The performances of learning-based fitting and optimization-based fitting are evaluated in Section~\ref{sec:bm_eval}, denoted as FaceScape(Learn) and FaceScape(Opti.) respectively.

\subsection{Displacement Map Prediction}
Detailed geometry is expressed by displacement maps for our predicted model.  In contrast to static details which are only related to the specific expression in a certain moment, dynamic detail expresses the geometry details in varying expressions.  Since the single displacement map cannot represent the dynamic details, we try to predict multiple displacement maps for 20 basic expressions in FaceScape using a deep neural network.

We observed that the displacement map in a certain expression could be decoupled into static components and dynamic components.  The static part tends to be static in different expressions and is mostly related to intrinsic features like pores, nevus, and organs.  The dynamic part varies in different expressions and is related to surface shrinking and stretching.  We use a deforming map to model the surface motion, which is defined as the difference of vertices' 3D position from source expression to target expression in the UV space.  As shown in Figure~\ref{fig:detail}, we can see the variance between displacement maps is strongly related to the deforming map, and the static features in displacement maps are related to the texture. So we feed motion maps and textures to a CNN to predict the displacement map for multiple expressions.

We use pix2pixHD\cite{wang2018high} as the backbone of our neural network to synthesize high-resolution displacement maps.  The input of the network is the stack of deforming maps and textures in UV space, which can be computed from the recovered base model.  Similar to \cite{wang2018high}, the combination of adversarial loss $L_{adv}$ and feature matching loss $L_{FM}$ is used to train our net with the loss function formulated as:
\begin{equation}
\begin{aligned}
L = \mathop{\min}_{G} ((&\mathop{\max}_{D_1, D_2, D_3} \sum_{k=1,2,3} L_{adv}(G,D_k)) \\& + \lambda \sum_{k=1,2,3}L_{FM}(G,D_k))
\end{aligned}
\end{equation}
where $G$ is the generator; $D_1, D_2$ and $D_3$ are discriminators that have the same LSGAN\cite{mao2017least} architecture but operate at different scales; $\lambda$ is the weight of feature matching loss, and the feature matching loss $L_{FM}$ is the same as defined in pix2pixHD~\cite{wang2018pix2pixHD}.

\subsection{Dynamic Detail Synthesis}
Inspired by\cite{nagano2018pagan}, we synthesize displacement map $F$ for an arbitrary expression corresponding to specific blendshape weight $\boldsymbol{\alpha}$, using a weighted linear combination of generated displacement maps $\hat{F}_0$ in neutral expression and $\hat{F}_i$ in other 19 key expressions:
\begin{equation}
\begin{aligned}
\mathop{F}=M_0 \odot \hat{F}_0 + \sum_{i=1}^{19} M_i \odot \hat{F}_i
\end{aligned}
\end{equation}
where $M$ is the weight mask with the pixel value between $0$ and $1$ , $\odot$ is element-wise multiplication operation. To calculate the weight mask, considering the blendshape expressions change locally, we first compute an activation mask $A_j$ in UV space for each blendshape mesh $e_j$ as:
\begin{equation}
\begin{aligned}
\mathop{A_j(p)}=||e_j(p)-e_0(p)||_2
\end{aligned}
\end{equation}
where $A_j(p)$ is the pixel value at position $p$ of the $j$th activation mask, $e_j(p)$ and $e_0(p)$ is the corresponding vertices position on blendshape mesh $e_j$ and neutral blendshape mesh $e_0$, respectively. The activation masks are further normalized between 0 and 1. Given the activation mask $A_j$ for each of the 51 blendshape meshes, the $i$th weight mask $M_i$ is formulated as a linear combination of the activation masks weighted by the current blendshape weight $\boldsymbol{\alpha}$ and fixed blendshape weight $\hat{\boldsymbol{\alpha}}_i$ corresponding to the $i$th key expression:
\begin{equation}
\begin{aligned}
\mathop{M_i}=\sum_{j=1}^{51}\boldsymbol{\alpha}^j\hat{\boldsymbol{\alpha}}_i^j{A_j}
\end{aligned}
\end{equation}
where $\boldsymbol{\alpha}^j$ is the $j$th element of $\boldsymbol{\alpha}$. $M_0$ is given by $M_0 = \max(0, 1-{\sum}_{i=1}^{19}M_i)$.

Many existing performance-driven facial animation methods generate blendshape weights with depth camera\cite{weise2011realtime, li2013realtime, bouaziz2013online} or single RGB camera\cite{cao20133d, cao2014displaced,chaudhuri2019joint}. As blendshape weights have semantic meaning, it's easy for artists to manually adjust the rigging parameters. 

\section{Experiments}

\begin{figure}[t]
\begin{center}
    \includegraphics[width=0.9\linewidth]{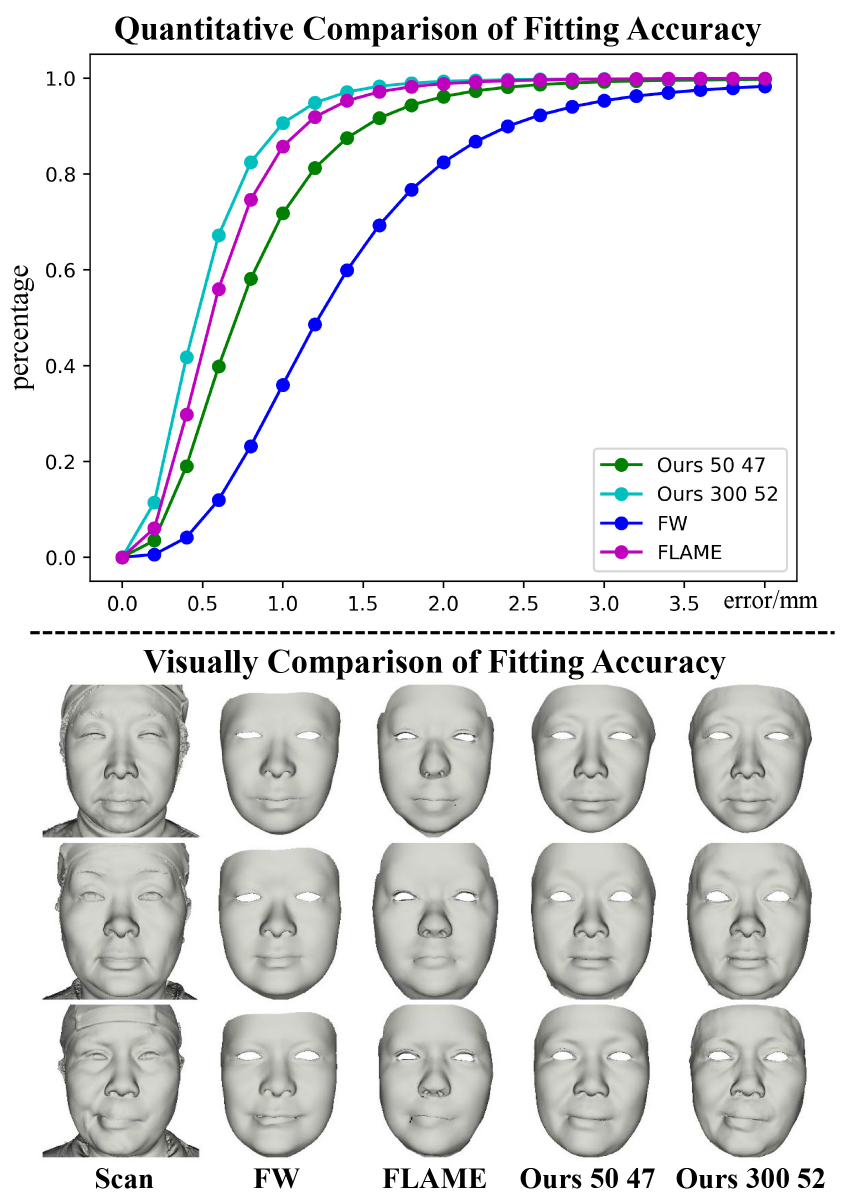}
\end{center}
    \vspace{-0.2in}
    \caption{Comparison of Reconstruction Error for parametric model generated by FaceScape and previous datasets.}
\label{fig:representation}
\end{figure}

\begin{figure*}[t]
\begin{center}
    \includegraphics[width=0.8\linewidth]{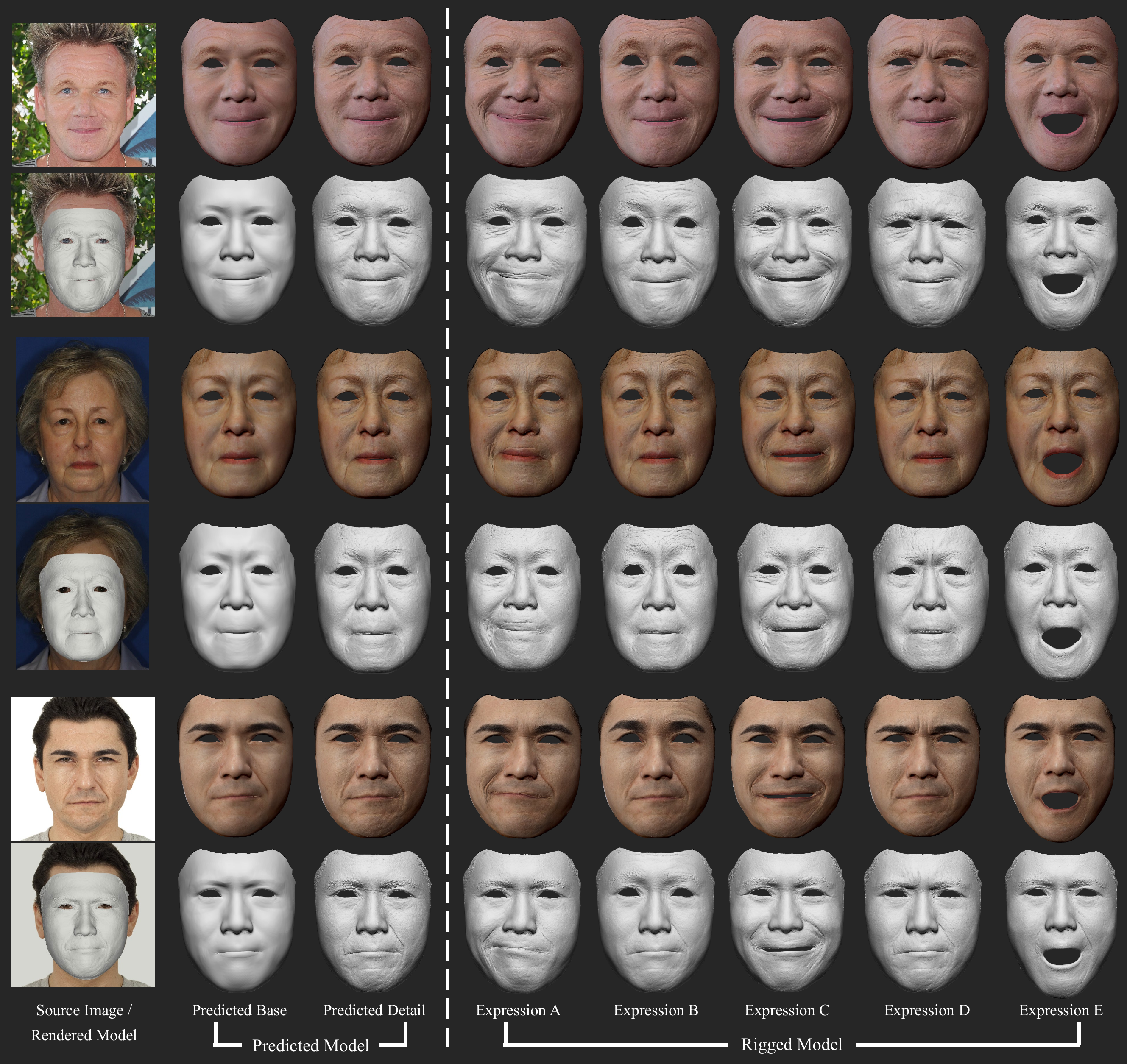}
\end{center}
    \vspace{-0.1in}
    \caption{We show our predicted faces in the source expression and rigged expressions.  It is worth noting that the wrinkles in rigged expressions are predicted from the source image.}
\label{fig:results}
\end{figure*}

\begin{figure}[t]
\begin{center}
    \includegraphics[width=1.0\linewidth]{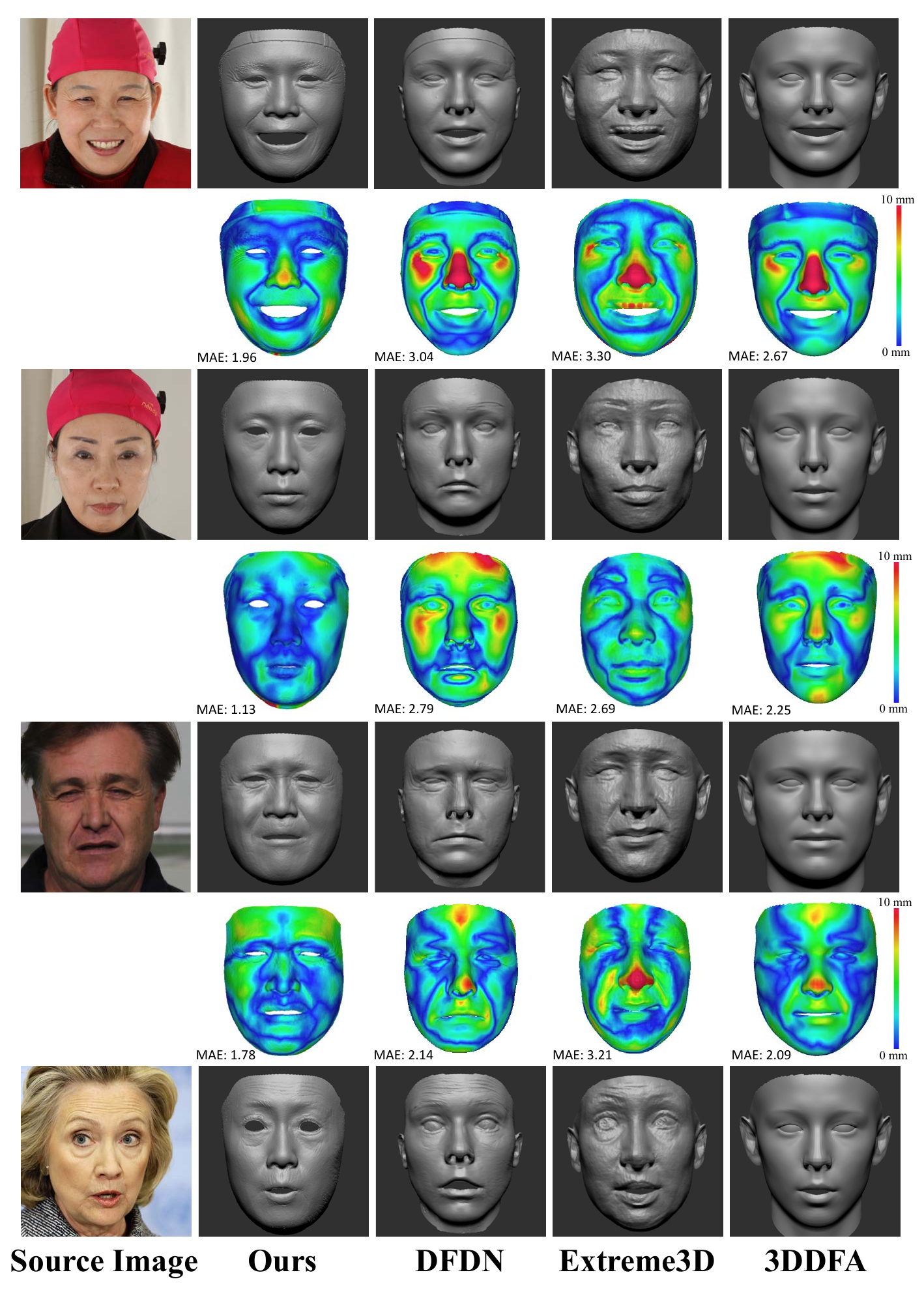}
\end{center}
    \vspace{-0.3in}
    \caption{Comparison of static 3D face prediction with previous methods.  The images in the top two rows are from FaceScape, the image in the third row is from Volker sequence\cite{valgaerts2012lightweight}, and the image in the bottom row is from the Internet.  The top three images are with ground truth shapes, so we evaluate the reconstruction error and show the heat map below each row.  
    }
\label{fig:compare}
\end{figure}

\subsection{Implementation Details}
\label{sec:implement_detail}

We use $847$ identities in our dataset as training data with a total of $16940$ displacement maps, leaving $50$ identities for the evaluation experiments and potential evaluation tasks in the future. We use the Adam optimizer to train the network with a learning rate as $2e^{-4}$. The input textures and output displacement maps' resolution of our network is both $1024 \times 1024$. We use $50$ identity parameters, $52$ expression parameters, and $100$ albedo parameters for our parametric model in all experiments.

\subsection{Evaluation of Representation Capability}
\label{sec:exp_represent}

We evaluate the representation capability of our model by fitting it to scanned 3D meshes not part of the training data. We compare our model to FaceWarehouse(FW)\cite{cao2013facewarehouse} and FLAME\cite{li2017learning} by fitting them to our self-captured test set, which consists of $1000$ high-quality meshes from $50$ testing subjects performing $20$ different expressions each. FW has $50$ identity parameters and $47$ expression parameters, so we use the same number of parameters for a fair comparison. To compare with FLAME which has $300$ identity parameters and $100$ expression parameters, we use $300$ identity parameters and all $52$ expression parameters. Figure~\ref{fig:representation} shows the cumulative reconstruction error. Our bilinear face model achieves much lower fitting error than FW using the same number of parameters and also outperforms FLAME using even fewer expression parameters. The visual comparison in Figure~\ref{fig:representation} shows our model could produce more mid-scale details than FW and FLAME, leading to more realistic fitting results.

\subsection{Evaluation of 3D Model Prediction}
\label{sec:eval_3d_pred}

We show the predicted riggable 3D faces in Figure~\ref{fig:results}, including a predicted 3D face at the source expression and rigged 3D faces at 5 expressions other than the source expression.
We can see that the predicted models contain photo-realistic detailed wrinkles, which deform as the expression changes. These are the dynamic details learned from the FaceScape dataset and can be synthesized from a single input image.
More rendered results including challenging cases, failure cases, and animations are shown in the supplementary material.

We compare our method with several previous methods qualitatively in Figure~\ref{fig:compare} and quantitatively in Table~\ref{tab:recon_error}. The test set of FaceScape and sequence~1 of Volker sequence~\cite{valgaerts2012lightweight} are used for the evaluation. Specifically, the data from the FaceScape test set contain 50 identities and each with 20 captured expressions, and the data from the Volker sequence contains $200$ continuous frames of a single identity. ``All Exp.'' means all 20 expressions are evaluated while ``Source Exp.'' means only the predicted results at the source expression are evaluated. Since the Volker sequence only provides ground truth for the source expression, we only evaluate the results of the source expression for the Volker sequence. The accuracy is measured with the mean absolute point-to-surface distance from the predicted mesh to the ground truth. As the back side of the 3D head models in the Volker sequence is inaccurate, we only extract the facial region from the head models for evaluation. It is worth noting that since the Volker sequence contains only one identity while Facescape contains multiple identities, the reported standard deviations (S.D.) in Table~\ref{tab:recon_error} of the former are significantly smaller than those of the latter. 

In Figure~\ref{fig:compare}, as most of the detailed faces predicted by previous works cannot be directly rigged to other expressions, we only show the face shape in the source expression. The accuracy is measured with the mean absolute point-to-surface distance from the predicted mesh to the ground truth.  Our results are visually better than previous methods, and also quantitatively better in the heat map of error. Table~\ref{tab:recon_error} also shows that our method leads in both the FaceScape test set and the Volker sequence. We consider the major reason for our method to perform the best in accuracy is the strong representation capability of our bilinear model, and the predicted details contribute to the visually plausible detailed geometry. As most of the faces in our testing set are facing the front, we would recommend referring to Section~\ref{sec:bm_eval} for a more comprehensive comparison with other methods.

\begin{figure}[t]
\begin{center}
    \includegraphics[width=1.0\linewidth]{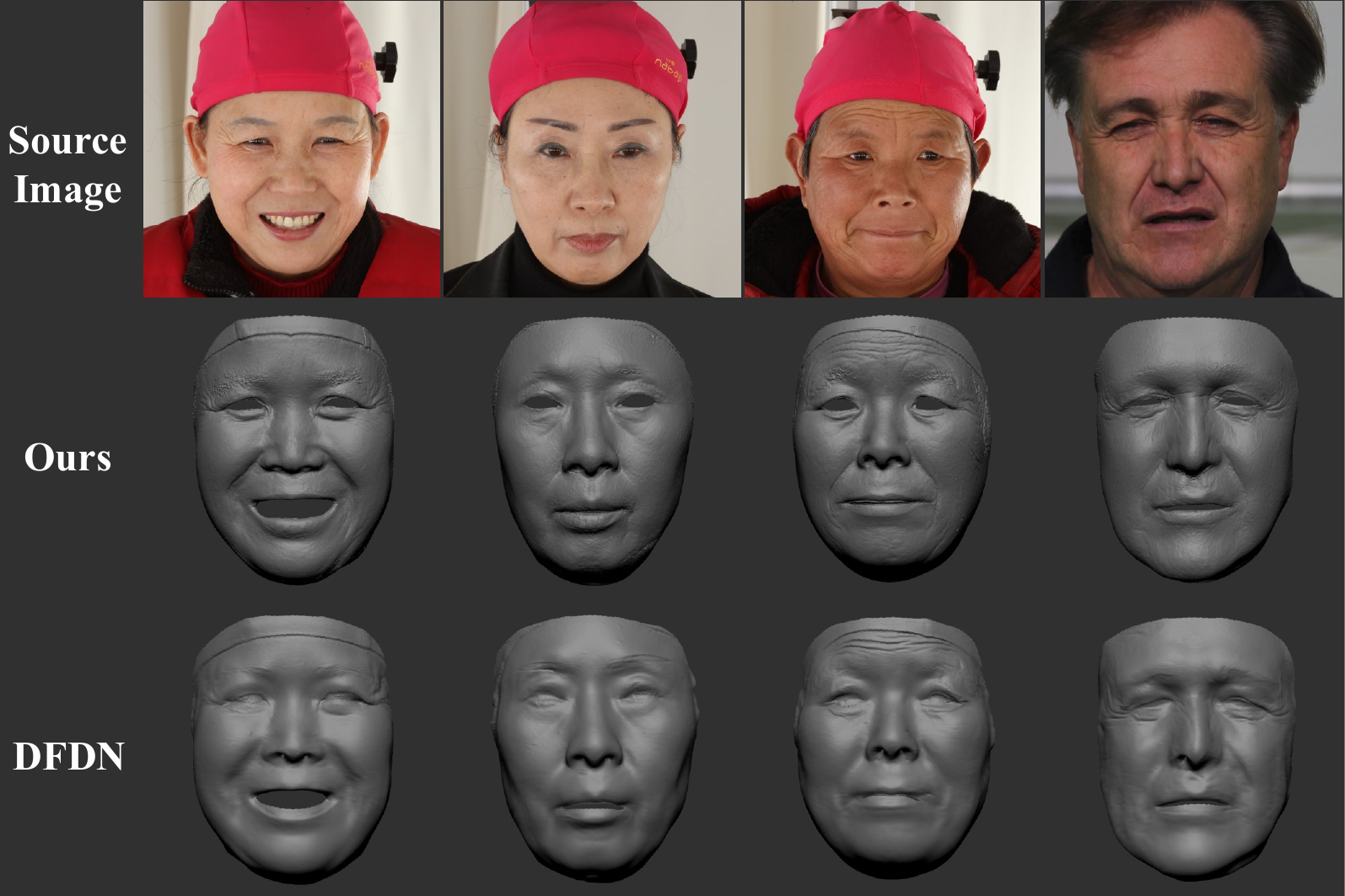}
\end{center}
    \vspace{-0.2in}
    \caption{Comparison of detail prediction. }
\label{fig:orthogonal}
\end{figure}

We also compare the detailed shape prediction with DFDN~\cite{chen2019photo} in Figure~\ref{fig:orthogonal}. In this experiment, NICP\cite{amberg2007optimal} is leveraged to register the base meshes of different methods to the ground truth, then the predicted detailed shape is visualized on the same base mesh. We can see that the detailed shape predicted by our method is more plausible.

\begin{table}[]
\center 
\caption{Quantitative Evaluation of 3D Model Prediction}
\vspace{-0.1in}
\begin{tabular}{@{}lcccc@{}}
\toprule
                                                               \multicolumn{1}{c}{Dataset~$\rightarrow$} & \multicolumn{2}{c}{FaceScape}                                   & \multicolumn{2}{c}{Volker Seq.} \\
\multicolumn{1}{c}{Method~$\downarrow$}                                     & Mean                           & S.D.                           & Mean             & S.D.              \\ \midrule
Our method (All Exp.)                                          & 1.39                           & 1.53                           & ---             & ---             \\ \midrule
Our method (Source Exp.)                                       & \textbf{1.22} & \textbf{1.08} & \textbf{2.54}             & \textbf{0.26}             \\
DFDN\cite{chen2019photo} (Source Exp.)        & 2.19                           & 1.79                           & 2.92             & 0.33             \\
Extreme3D\cite{tran2018extreme} (Source Exp.) & 2.06                           & 1.60                           & 4.04             & 0.47             \\
3DDFA\cite{zhu2017face} (Source Exp.)         & 2.17                           & 1.80                           & 2.76             & 0.29             \\ \bottomrule
\end{tabular}
\label{tab:recon_error}
\end{table}

\subsection{Ablation Study}
\begin{figure}[t]
\begin{center}
    \includegraphics[width=1.0\linewidth]{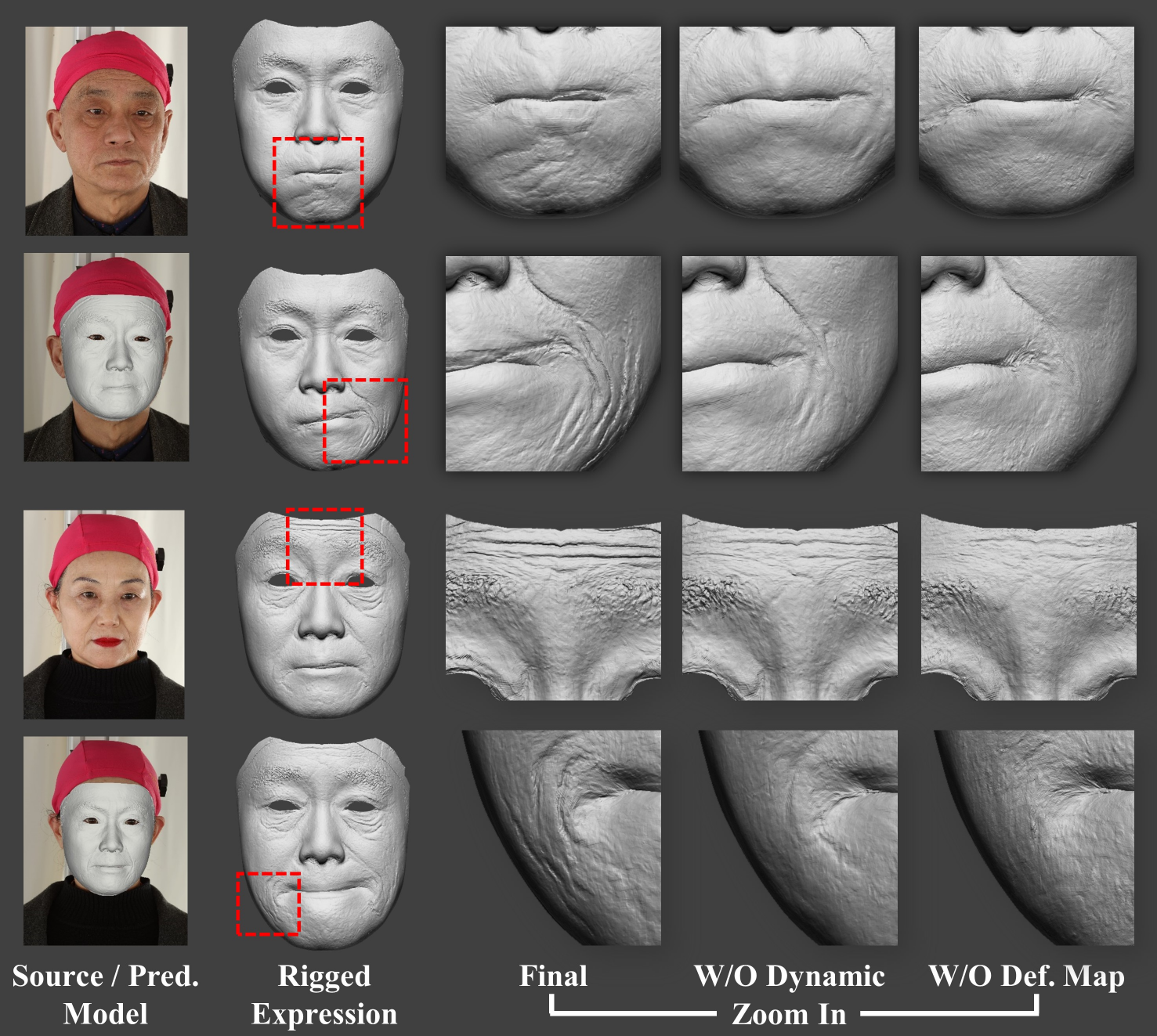}
\end{center}
    \vspace{-0.2in}
    \caption{Ablation study.  Our final model can recover wrinkles in rigged expressions, while the method W/O deforming map and W/O dynamic details cannot.}
\label{fig:ablation}
\end{figure}

\textbf{W/O dynamic detail.}  We try to use only one displacement map from the source image for rigged expressions, and the other parts remain the same.  As shown in Figure~\ref{fig:ablation}, we find that the rigged model with dynamic detail shows the wrinkles caused by various expressions, which are not found in the W/O dynamic setting.

\noindent\textbf{W/O deforming Map.}  We change the input of our displacement map prediction network by replacing the deforming map with one-hot encoding for each of the $20$ target expressions.  As shown in Figure~\ref{fig:ablation}, we find the results without deforming map (W/O Def. Map) contain few details caused by expressions.

\subsection{Limitations}

We can see some limitations of our SVFR method through experiments and analysis.

\noindent\textbf{Degraded performance for side views.} The quantitative evaluations in Section~\ref{sec:bm_eval} and the visualized results in the supplementary material show that our results of a side view are worse than that of a frontal view. Though our base mesh fitting is robust to side-view input, the displacement map prediction cannot handle the self-occlusion regions well, which leads to some artifacts for the occluded face. The reason is that our network lacks the capability to hallucinate unseen textures, and the wrongly predicted texture maps lead to wrongly predicted displacement maps.

\noindent\textbf{Degraded performance for occlusion and extreme lighting.} The visualized results in the supplementary material also show that our performance is degraded in some extreme lighting conditions and for partially occluded faces. The reason is that our network lacks the capability to recover the texture from the extremely dark image and hallucinate occluded textures, and the degraded predicted texture maps lead to inaccurate displacement map prediction. 

\noindent\textbf{Racial bias.}
We observed that our method struggles to reconstruct the appearance features of some non-Asian races, as analyzed in failure cases in the supplementary material. We think the reason is that our parametric model of facial geometry and textures is built mostly upon Asian faces. In the future, with more releases of high-quality 3D face data of various races, racial bias can be eliminated by merging multiple data sets.

\section{Benchmark}

In recent years, a large number of methods are presented to recover 3D facial shapes from a single image. Surprisingly, there is very few benchmark and data for single-view face reconstruction that can take all aspects into account. The common evaluating way is to run the method on tuples of image-model pairs, where the image is the input, then the error between the output and the ground-truth model is reported.  These image-model pairs for evaluation should meet the following requirements: 1) Accurate - the 3D model is accurately captured and is well aligned to the image; 2) Photo-realistic - the image should be like the in-the-wild photos; 3) Amount - The amount of data should be large enough to cover various appearance, poses, expressions, environments, and lightings. 
In this section, we propose the FaceScape benchmark to evaluate the accuracy of 3D facial shape reconstruction, consisting of FaceScape-Wild (FS-Wild) and FaceScape-Lab (FS-Lab) data.  FS-Wild data takes advantage of face-swapping methods to synthesize photo-realistic in-the-wild images with exact ground-truth 3D shapes.  FS-Lab dataset uses the unpublished in-the-lab images captured by our multi-view system and the corresponding accurate 3D model as ground truth. We will explain the data generation, metrics, and evaluation results in the following subsections.

\subsection{FS-Wild data generation}

\begin{figure*}[t]
\begin{center}
\includegraphics[width=1.0\linewidth]{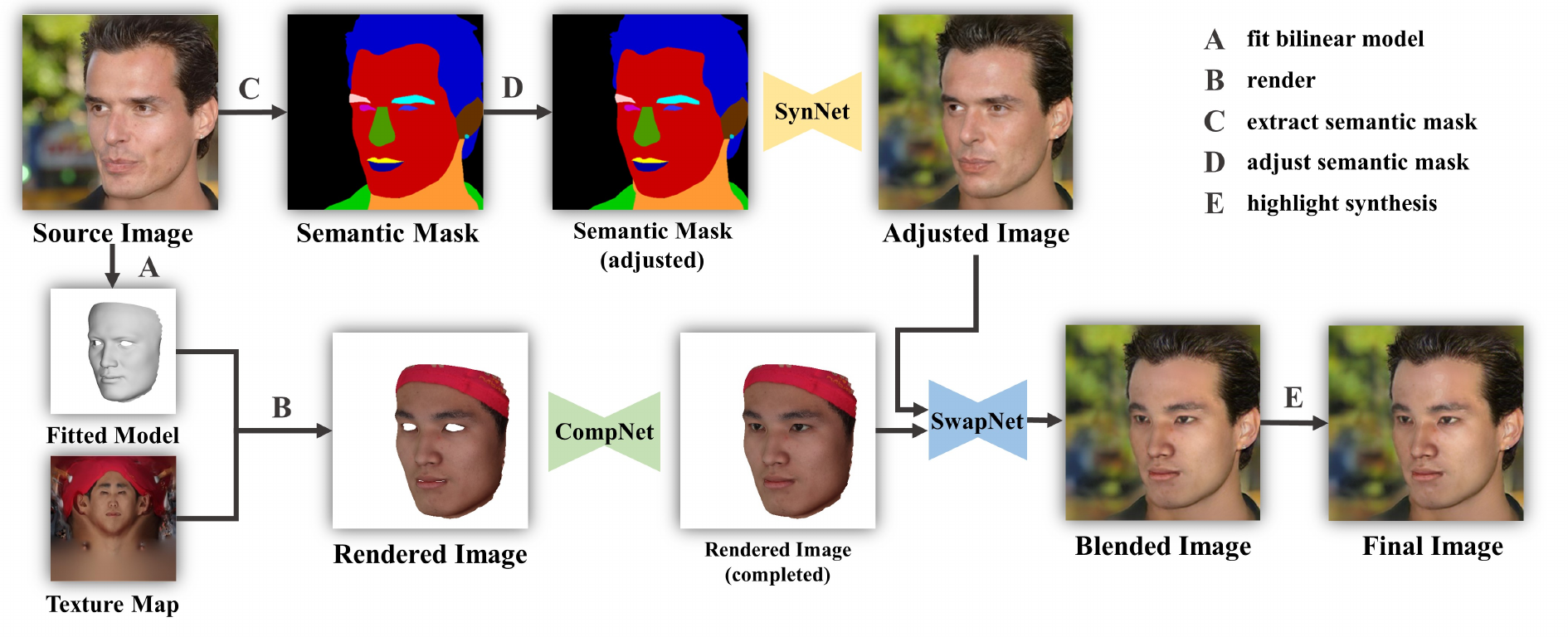}
\vspace{-0.3in}
\end{center}
\caption{Overview of the proposed in-the-wild face rendering approach.}
\label{fig:render_pipeline_v1}
\end{figure*}

We create FS-Wild data by swapping the in-the-wild faces with rendered faces from ground-truth models. In this way, we obtain the photo-realistic in-the-wild facial images and the corresponding exact 3D ground-truth shape for evaluation. The pipeline to synthesize the final image is shown in Figure~\ref{fig:render_pipeline_v1}, which consists of three stages. 

In the first stage (A/B/CompNet in Figure~\ref{fig:render_pipeline_v1}), we fit the bilinear model to the source in-the-wild image using the algorithm described in Section~\ref{sec:bmfit}. A texture map is then randomly selected from the pool of TU models, according to the gender and age of the subject in the source image.  The fitted model and texture map are then rendered to the image with lighting in the source image, which is recovered by Deng~\etal's method~\cite{deng2019accurate} in the form of the Spherical Harmonics (SH) model~\cite{sloan2002precomputed, ramamoorthi2006modeling}.
Considering that the eyes and mouth are missing, we train a CompNet to complement the eyes and mouth. CompNet adopts pix2pixHD\cite{wang2018pix2pixHD} as the backbone and is trained using the captured images with eyes and mouth.  We can see that CompNet generates photo-realistic eyes and mouth for the rendered image, and we also obtain the fitted model as the ground-truth shape.

In the second stage (C/D/SynNet in Figure~\ref{fig:render_pipeline_v1}), we first extract the semantic mask, then adjust the facial mask to fit the completed image.  For the images from CelebAMask-HQ\cite{lee2020maskgan}, the provided ground-truth masks are used. For the images from AFLW\cite{koestinger11a} that do not contain a ground-truth mask, BiSeNet\cite{yu2018bisenet} is used to extract the semantic mask from the source image.  The adjustment aims to make the facial mask of the source image and rendered image completely aligned. As model fitting cannot make the rendered mask exactly align with the facial mask, mask adjustment is essential for the following face-swap stage. Otherwise, the face-swapping result will contain artifacts around the cheek. Then, SynNet is used to generate photo-realistic faces from the adjusted semantic masks. We refer to SEAN~\cite{zhu2020sean} as our SynNet, which is a generative adversarial network conditioned on a semantic mask. In this way, the photo-realistic image that is exactly aligned with the fitted model is generated.

In the last stage (SwapNet/E in Figure~\ref{fig:render_pipeline_v1}), we use SwapNet to replace the facial region of the adjusted image with that of the completed image. There are a large number of works that study the face-swapping problem, but most of them cannot maintain the 3D structure of the completed images. By experiments, we find the inpainting and blending module in FSGAN\cite{nirkin2019fsgan} meets our requirement, which keeps a faithful 3D structure of the completed image while synthesizing photo-realistic in-the-wild images.  Finally, we obtained the final image as the input for evaluation, and the fitted model as the corresponding ground-truth shape. 
We observed that the rendering results with the spherical harmonic lighting model and after SwapNet lack specular highlights, which makes the rendering results unrealistic. Therefore, we leverage a highlight synthesis module to supplement specular highlights on the face. Inspired by Hou~\etal's work~\cite{hou2022face}, we first render the ground-truth facial mesh model with a skin-like material under Phong shading~\cite{phong1975illumination}, then transfer the shadows and specular highlights from the rendered face to the result of the SwapNet. The lighting directions are extracted from the spherical harmonic lighting model estimated in stage B, while the lighting intensity and the specular reflections are randomly selected. 

\textbf{Implementation details.}
The backbones of our CompNet is the pix2pixHD\cite{wang2018pix2pixHD} with a U-Net as the generator. The CompNet is trained with reconstruction loss $\mathcal{L}_{rec}$ and adversarial loss $\mathcal{L}_{adv}$:\\

\begin{equation}
\mathop{\min}_{G} ((\mathop{\max}_{D} \mathcal{L}_{adv}(G,D)) + \lambda \mathcal{L}_{rec}(G)),
\end{equation}
where $G$ is the generator, $D$ is the discriminator and $\lambda$ is the weight of reconstruction loss.
The reconstruction loss $\mathcal{L}_{rec}$ is given by:
\begin{equation}
\mathcal{L}_{rec} = \left \| {G(x)}-{x} \right \|_1.
\end{equation}
The adversarial loss is defined as:
\begin{equation}
    \mathcal{L}_{adv}=\mathbb{E}_{x,y}[\log D(x,y)]+\mathbb{E}_{x}[\log (1-D(x,G(x)))].
\end{equation}

Our SynNet is trained on CelebAMask-HQ following the training methods in SEAN~\cite{zhu2020sean}. To adjust the semantic mask, we first remove the headgear region from the facial region of the rendered image, then take the intersection of the facial part of the semantic mask and the rendered image. For the region that is contained by the semantic mask but is not contained by the rendered image, the semantic label is replaced by the nearest non-facial label. In this way, the adjusted mask has the same face contour as the rendered image.
The input image size of SynNet and CompNet is $512 \times 512$ and the output image size of SwapNet is $256 \times 256$.

\textbf{Results of In-the-wild Face Rendering.} The benchmark of FS-Wild consists of 400 face images of 400 synthesized subjects. The data are uniformly divided into $4$ sets according to the angle between camera orientation and face orientation ($0-5^\circ$, $5^\circ-30^\circ$, $30^\circ-60^\circ$, $60^\circ-90^\circ$), with a reference 3D face model per subject. The images consist of indoor and outdoor images, neutral expression and expressive face images, and varying viewing angles ranging from frontal view to side view. The samples of FS-Wild data are shown in the supplementary material.

\subsection{FS-Lab data generation}

We render $330$ images using the $20$ detailed 3D models, which are randomly selected from the unpublished testing set of FaceScape. These subjects' age ranges from $17$ to $63$, with an average age of $38.7$.  Centering on the head and starting from the front, we select $11$ different camera locations with (yaw, pitch) coordinates:

\noindent\textbf{$\bullet$ $1$ camera at exact front}: $(0^\circ, 0^\circ)$.

\noindent\textbf{$\bullet$ $8$ cameras deflecting $30^\circ$}: 

$(0^\circ, \pm30^\circ)$, $(\pm21.47^\circ, \pm21.47^\circ)$, $(\pm30^\circ, 0^\circ)$.

\noindent\textbf{$\bullet$ $2$ cameras deflecting $60^\circ$}: $(\pm60^\circ, 0^\circ)$.

\begin{figure}[t]
\begin{center}
    \includegraphics[width=1.0\linewidth]{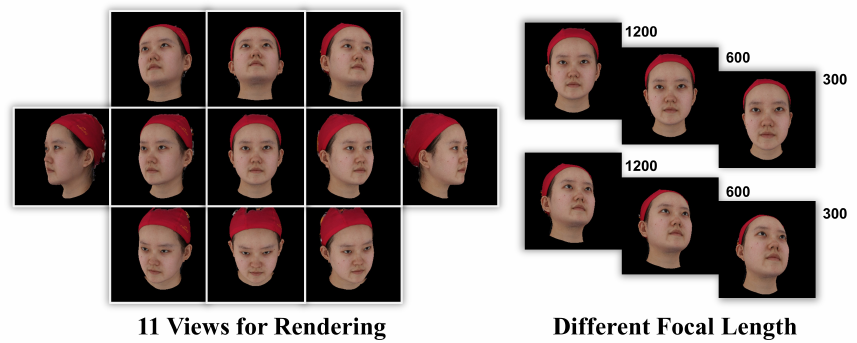}
\end{center}
    \vspace{-0.2in}
    \caption{Samples of in-the-lab data in different views and different focal lengths.}
\label{fig:lab_data}
\end{figure}

\noindent The samples images for $11$ camera locations are shown in Figure~\ref{fig:lab_data}.

The images are rendered in the resolution of $256 \times 256$ using $3$ different focal lengths: $1200$ (long focal), $600$ (middle focal), $300$ (short focal). The samples for different focal lengths are shown in Figure~\ref{fig:lab_data}.

\begin{table*}[]
\centering
\caption{Quantitative evaluation on FS-Wild dataset categorized by pose angle.}
\vspace{-0.1in}
\begin{tabular}{lccccccccccccc}
\toprule
                            \multicolumn{1}{c}{Pose Angle $\rightarrow$}
                                       & 
                            \multicolumn{3}{c}{$0^\circ-5^\circ$}             & \multicolumn{3}{c}{$5^\circ-30^\circ$}            & \multicolumn{3}{c}{$30^\circ-60^\circ$}            & \multicolumn{3}{c}{$60^\circ-90^\circ$}             &           \\
\multicolumn{1}{c|}{Method \ $\downarrow$} & CD   & MNE   & \multicolumn{1}{c|}{CR} & CD   & MNE   & \multicolumn{1}{c|}{CR} & CD   & MNE   & \multicolumn{1}{c|}{CR} & CD    & MNE   & \multicolumn{1}{c|}{CR} & Success Rate \\ \midrule
Ext3dFace\cite{tran2018extreme}   & 4.96    & 0.155     & 61.7    & 5.58    & 0.174     & 56.1    & 7.55    & 0.205     & 40.9    & 26.03   & 0.275     & 27.0    & 85.3      \\
PRNet\cite{feng2018joint}           & 2.64    & 0.116     & 83.3    & 3.07    & 0.113     & 82.9    & 4.28    & 0.118     & 78.6    & 3.88    & 0.141     & 75.2    & 100.0     \\
Deep3DFaceRec\cite{deng2019accurate}   & 3.21    & 0.096     & 78.0    & 4.13    & 0.101     & 77.4    & 5.74    & 0.113    & 71.3    & 8.91    & 0.168     & 57.3    & 100.0     \\
RingNet\cite{sanyal2019learning}         & 2.41 & 0.084 & 99.8 & 2.98 & 0.085 & 99.7 & 4.86 & 0.100 & 98.4 & 10.79 & 0.190 & 96.8 & 100.0 \\
DFDN\cite{chen2019photo}            & 3.65 & 0.091 & 86.9 & 3.30 & 0.092 & 86.7 & 7.28 & 0.128 & 84.7 & 27.03 & 0.302 & 57.0 & 88.8 \\
DF2Net\cite{zeng2019df2net}          & 2.96 & 0.116 & 58.3 & 4.17 & 0.126 & 56.2 & 6.73 & 0.158 & 46.6 & 21.27 & 0.344 & 25.8 & 74.9 \\
UDL\cite{chen2020self}             & 2.25 & 0.084 & 69.1 & 2.50 & 0.088 & 68.2 & 3.38 & 0.101 & 65.1 & 6.31 & 0.176 & 49.0 & 87.0 \\
FaceScape(Opti.)\cite{yang2020facescape} & 1.97 & 0.080 & 85.0 & 2.77 & 0.089 & 82.7 & 3.92 & 0.107 & 77.9 & 7.11 & 0.169 & 66.7 & 95.5 \\
FaceScape(Learn)\cite{yang2020facescape} & 2.63 & 0.085 & 86.9 & 3.62 & 0.092 & 86.5 & 4.11 & 0.099 & 85.3 & 9.09 & 0.151 & 70.8 & 100.0 \\
MGCNet\cite{shang2020self}          & 3.11 & 0.072 & 84.5 & 2.92 & 0.072 & 84.6 & 2.85 & 0.070 & 81.6 & 4.20 & 0.091 & 74.4 & 100.0 \\
3DDFA\_v2\cite{guo2020towards}      & 2.56 & 0.075 & 86.3 & 2.69 & 0.075 & 86.0 & 3.22 & 0.080 & 83.1 & 3.68 & 0.093 & 80.0 & 100.0 \\
SADRNet\cite{ruan2021sadrnet}         & 6.69 & 0.114 & 90.4 & 6.87 & 0.113 & 89.6 & 6.41 & 0.102 & 84.5 & 8.69 & 0.163 & 82.7 & 100.0 \\
LAP\cite{zhang2021learning}             & 4.07 & 0.107 & 92.8 & 4.40 & 0.116 & 91.9 & 6.06 & 0.147 & 87.2 & 13.60 & 0.203 & 68.5 & 100.0 \\
DECA\cite{feng2021learning}            & 2.92 & 0.079 & 99.9 & 2.65 & 0.078 & 99.9 & 2.91 & 0.080 & 99.7 & 4.81 & 0.116 & 99.7 & 100.0 \\
\bottomrule
\end{tabular}
\label{tab:eval_wild_a}
\end{table*}

\begin{table*}[]
\centering
\caption{Quantitative evaluation on FS-Lab benchmark categorized by pose angle.}
\vspace{-0.1in}
\begin{tabular}{lcccccccccc}
\toprule
                            \multicolumn{1}{c}{Pose Angle $\rightarrow$}  &      & $0^\circ$ &                         &      & $30^\circ$ &                         &       & $60^\circ$ &                         &           \\
\multicolumn{1}{c|}{Method \ $\downarrow$} & CD   & MNE    & \multicolumn{1}{c|}{CR} & CD   & MNE     & \multicolumn{1}{c|}{CR} & CD    & MNE     & \multicolumn{1}{c|}{CR} & Success Rate \\ \midrule
Ext3dFace\cite{tran2018extreme}   & 4.59    & 0.131     & 86.2    & 7.42    & 0.170     & 69.1    & 8.51    & 0.175     & 55.2    & 85.9      \\
PRNet\cite{feng2018joint}           & 2.94    & 0.133     & 92.5    & 3.40    & 0.125     & 90.1    & 3.74    & 0.122     & 85.2    & 100.0     \\
Deep3DFaceRec\cite{deng2019accurate}   & 3.99    & 0.106     & 87.6    & 5.90    & 0.120     & 81.3    & 5.55    & 0.137     & 75.3    & 98.9      \\
RingNet\cite{sanyal2019learning}         & 3.62    & 0.102     & 99.9    & 5.03    & 0.111     & 99.7    & 6.82    & 0.151     & 94.5    & 100.0     \\
DFDN\cite{zeng2019df2net}            & 4.28    & 0.111     & 98.4    & 6.71    & 0.132     & 95.2    & 23.63   & 0.280     & 81.0    & 94.7      \\
DF2Net\cite{zeng2019df2net}          & 4.48    & 0.152     & 64.1    & 7.64    & 0.200     & 52.2    & -1.00   & -1.000    & -100.0  & 73.6      \\
UDL\cite{chen2020self}             & 2.21    & 0.092     & 79.5    & 5.34    & 0.123     & 71.3    & 5.63    & 0.167     & 61.9    & 87.0      \\
FaceScape(Opti.)\cite{yang2020facescape} & 3.21    & 0.090     & 94.2    & 4.87    & 0.119     & 86.2    & 4.68    & 0.146     & 81.7    & 92.0      \\
FaceScape(Learn)\cite{yang2020facescape} & 2.40    & 0.086     & 96.7    & 7.28    & 0.124     & 87.7    & 3.87    & 0.108     & 90.5    & 100.0     \\
MGCNet\cite{shang2020self}          & 3.45    & 0.085     & 92.7    & 3.91    & 0.093     & 90.1    & 3.65    & 0.090     & 83.2    & 100.0     \\
3DDFA\_v2\cite{guo2020towards}       & 3.05    & 0.093     & 95.2    & 3.41    & 0.096     & 93.8    & 3.82    & 0.097     & 88.2    & 100.0     \\
SADRNet\cite{ruan2021sadrnet}         & 4.25    & 0.109     & 95.8    & 7.07    & 0.137     & 94.9    & 7.09    & 0.148     & 87.6    & 100.0     \\
LAP\cite{zhang2021learning}             & 4.27    & 0.112     & 96.4    & 7.33    & 0.149     & 93.2    & 8.70    & 0.195     & 85.6    & 99.2      \\
DECA\cite{feng2021learning}            & 3.30    & 0.093     & 99.8    & 4.14    & 0.100     & 99.4    & 4.20    & 0.107     & 97.1    & 100.0     \\ \bottomrule
\end{tabular}
\label{tab:eval_lab_a}
\end{table*}

\begin{table*}[]
\centering
\caption{Quantitative evaluation on FS-Lab benchmark categorized by focal length.}
\vspace{-0.1in}
\begin{tabular}{lcccccccccc}
\toprule
                            \multicolumn{1}{c}{Focal Length $\rightarrow$}   &      & Long$(1200)$ &                         &      & Mid$(600)$ &                         &       & Short$(300)$ &                         &           \\
\multicolumn{1}{c|}{Method \ $\downarrow$} & CD   & MNE    & \multicolumn{1}{c|}{CR} & CD   & MNE   & \multicolumn{1}{c|}{CR} & CD    & MNE   & \multicolumn{1}{c|}{CR} & Success Rate \\ \midrule
Ext3dFace\cite{tran2018extreme}   & 7.25    & 0.167     & 69.4    & 6.72    & 0.162     & 64.9    & 6.03    & 0.160     & 61.4    & 85.9      \\
PRNet\cite{feng2018joint}           & 3.42    & 0.125     & 89.4    & 3.48    & 0.124     & 89.0    & 3.79    & 0.128     & 90.2    & 100.0     \\
Deep3DFaceRec\cite{deng2019accurate}   & 5.67    & 0.122     & 80.8    & 5.28    & 0.117     & 79.2    & 4.90    & 0.114     & 81.1    & 98.9      \\
RingNet\cite{sanyal2019learning}         & 5.23    & 0.117     & 98.8    & 5.25    & 0.117     & 99.4    & 5.37    & 0.119     & 99.8    & 100.0     \\
DFDN\cite{chen2019photo}            & 9.05    & 0.153     & 93.3    & 9.40    & 0.149     & 92.8    & 9.30    & 0.146     & 94.6    & 94.7      \\
DF2Net\cite{zeng2019df2net}          & 7.26    & 0.194     & 53.7    & 6.97    & 0.191     & 51.2    & 6.39    & 0.183     & 49.5    & 73.6      \\
UDL\cite{chen2020self}             & 5.06    & 0.126     & 70.9    & 4.91    & 0.124     & 69.2    & 4.95    & 0.125     & 69.7    & 87.0      \\
FaceScape(Opti.)\cite{yang2020facescape} & 4.69    & 0.120     & 86.4    & 4.77    & 0.121     & 85.2    & 5.47    & 0.126     & 83.6    & 92.0      \\
FaceScape(Learn)\cite{yang2020facescape} & 6.21    & 0.118     & 89.0    & 6.19    & 0.118     & 88.8    & 6.43    & 0.125     & 86.4    & 100.0     \\
MGCNet\cite{shang2020self}          & 3.82    & 0.091     & 89.1    & 4.01    & 0.091     & 89.4    & 4.18    & 0.098     & 91.3    & 100.0     \\
3DDFA\_v2\cite{guo2020towards}       & 3.45    & 0.096     & 92.9    & 3.51    & 0.094     & 92.7    & 3.85    & 0.097     & 93.5    & 100.0     \\
SADRNet\cite{ruan2021sadrnet}         & 6.81    & 0.137     & 93.6    & 6.82    & 0.132     & 95.0    & 6.60    & 0.131     & 97.1    & 100.0     \\
LAP\cite{zhang2021learning}             & 7.30    & 0.154     & 92.1    & 7.06    & 0.151     & 91.4    & 6.75    & 0.150     & 91.7    & 99.2      \\
DECA\cite{feng2021learning}            & 4.07    & 0.100     & 99.0    & 4.19    & 0.101     & 99.6    & 5.81    & 0.122     & 99.8    & 100.0     \\ \bottomrule
\end{tabular}
\label{tab:eval_lab_f}
\end{table*}

\begin{figure*}[t]
\begin{center}
    \includegraphics[width=1.0\linewidth]{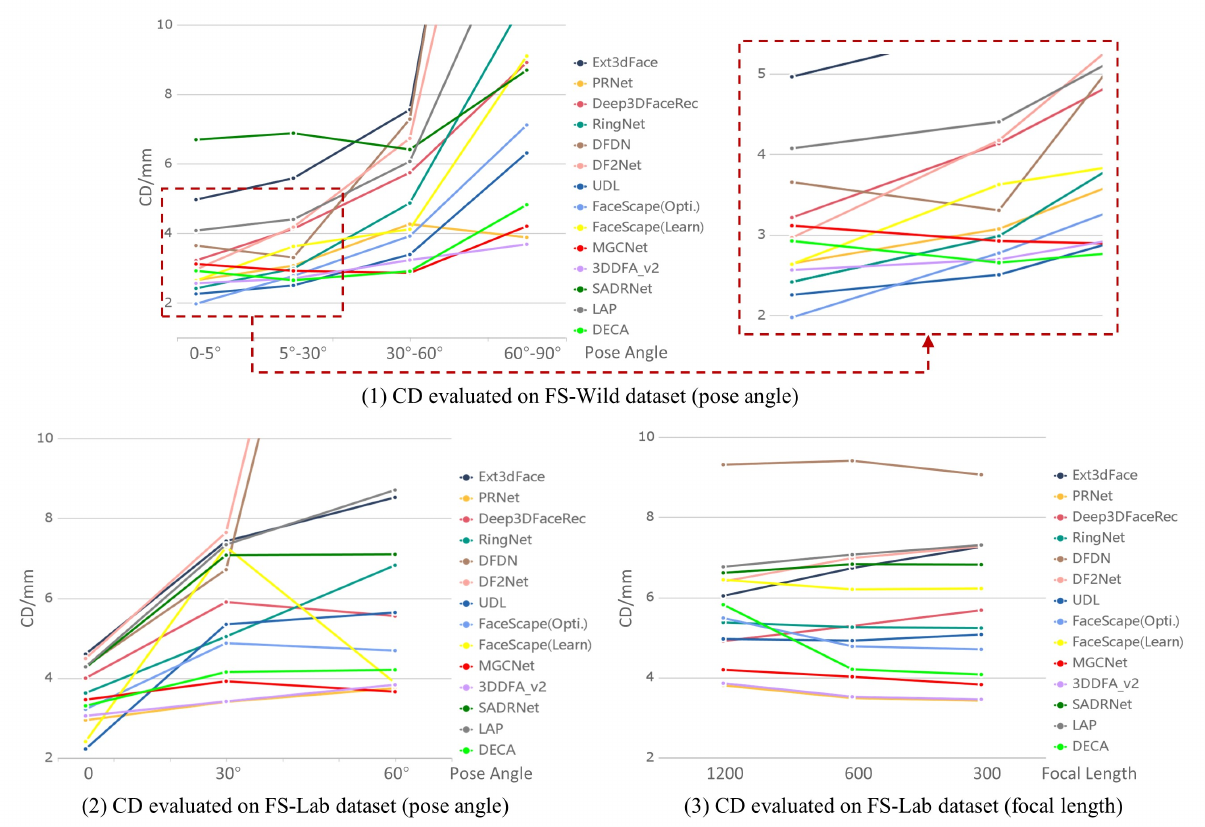}
\end{center}
    \vspace{-0.15in}
    \caption{Charts to visualize the quantitative evaluation. }
\label{fig:benchmark_chart}
\end{figure*}

\subsection{Metric}

We propose to quantitatively evaluate the accuracy of the single-view face reconstruction methods using Chamfer Distance (CD), Mean Normal Error (MNE), and Complete Rate (CR). Considering that some methods may fail on certain images, these metrics are only computed from the valid outputs, and we additionally report the success rate for the whole dataset.

\textbf{Pre-process.}
Firstly, we transform the predicted mesh and the ground-truth mesh into the camera coordinate. Considering different methods may use perspective projection cameras or orthogonal projection cameras, these projection models need to be adjusted to match the projection model used for ground-truth rendering. For FS-Wild data, all predicted meshes are transformed into orthogonal camera coordinates, and for FS-Lab data, all predicted meshes are transformed into the perspective camera coordinate with the ground-truth focal length. Then the depth of the predicted mesh is optimized to fit the ground-truth mesh by minimizing the depth difference. It is worth noting that we do not use ICP\cite{arun1987least} or 7-point alignment\cite{feng2018evaluation} to register the predicted mesh to the ground-truth mesh, as used in Section~\ref{sec:eval_3d_pred}, but only align them by moving the predicted mesh along the depth direction. We believe our alignment tells more because ICP and 7-point alignment changes the pose of the predicted mesh. Because of this, the chamfer distance reported in the benchmark is slightly larger than the error distance reported in Section~\ref{sec:eval_3d_pred}.

We observed that the grid density of the mesh is not uniform, which may cause bias in the quantitative evaluation. Therefore, all the predicted mesh and the ground-truth mesh are re-sampled by projecting them to the cylindrical coordinate. 
The central axis of the cylindrical coordinate crosses the barycenter of the ground-truth head mesh, pointing to the overhead direction. We found that projecting to a cylindrical coordinate is better than a sphere coordinate and Euclidean coordinate, because it causes the fewest occlusions and makes the sampling of points most uniform.  When re-sampling meshes from the position map in the sphere coordinate, the faces with edges larger than $15mm$ are ignored. 

\textbf{Chamfer Distance.} CD measures the overall error distance.  Given the processed predicted mesh $\mathcal{M}_{p}$ and the ground-truth mesh $\mathcal{M}_{g}$, chamfer distance is formulated as:

\begin{equation}
\begin{aligned}
{CD}({\mathcal{M}_{p}}, {\mathcal{M}_{g}}) = \frac{1}{N_{p}} \sum\limits_{x\in{\mathcal{M}_{p}}}^{N_{p}} \sum\limits_{y\in{\mathcal{M}_{g}}}^{N_{g}} {\min \left\| {x - y} \right\|}_2 + \\ \frac{1}{N_{g}} \sum\limits_{y\in{\mathcal{M}_{g}}}^{N_{g}} \sum\limits_{x\in{\mathcal{M}_{p}}}^{N_{p}} {\min \left\| {x - y} \right\|}_2
\end{aligned}
\end{equation}

where $N_{p}$, $N_{g}$ are the numbers of the vertices of the predicted mesh and the ground-truth mesh respectively. 

\textbf{Mean Normal Error (MNE).} MNE measures the accuracy of the detailed geometry in the middle-scale and small-scale. To compute MNE, we first render the predicted mesh and the ground-truth mesh in the cylindrical coordinate, generating predicted normal map $\mathcal{N}_p$ and ground-truth normal map $\mathcal{N}_g$. Only the intersection of the valid region for $\mathcal{N}_p$ and $\mathcal{N}_g$ are reserved, so the pixels in $\mathcal{N}_p$ and $\mathcal{N}_g$ are one-to-one matched. Then, MNE is formulated as:

\begin{equation}
\begin{aligned}
MNE(\mathcal{N}_p, \mathcal{N}_g) = \frac{1}{{N_n}}\sum\limits_{x,y \in \mathcal{N}_p, \mathcal{N}_g}^{N_n} {\left( {\frac{{x \cdot y}}{{\left\| x \right\|_2 \cdot \left\| y \right\|_2}}} \right)}
\end{aligned}
\end{equation}

where $N_n$ is the number of valid pixels in the two normal maps. 

\textbf{Complete Rate (CR).} CR measures the integrity of the reconstruction results. To compute CR, we first render the predicted mesh and the ground-truth mesh in the cylindrical coordinate as the position map $P_p$ and $P_g$ respectively. 
Then CR is formulated as:

\begin{equation}
\begin{aligned}
\eta  = \frac{{S(P_p \cap P_g)}}{{S(P_g)}}
\end{aligned}
\end{equation}

where $S(P)$ is the function that returns the area of the position map $P$.

\subsection{Evaluation}
\label{sec:bm_eval}

We evaluate 14 recent single-view face reconstruction methods, and the quantitative results are reported in Table~\ref{tab:eval_wild_a}~\ref{tab:eval_lab_a}~\ref{tab:eval_lab_f}. Table~\ref{tab:eval_wild_a} show the scores on FS-Wild data categorized by pose angle. Table~\ref{tab:eval_lab_a} and Table~\ref{tab:eval_lab_f} show the scores on FS-Lab data categorized by pose angle and focal length respectively.

It is worth noting that some methods are special in certain aspects:
FaceScape(Opti./Learn)\cite{yang2020facescape} and DECA\cite{feng2021learning} predicts riggable mesh model; LAP\cite{zhang2021learning} is trained with the in-the-wild photo collections, and MGCNet\cite{shang2020self} is trained with multi-view images; Ext3dFace\cite{tran2018extreme}, DFDN\cite{chen2019photo}, and DF2Net\cite{zeng2019df2net} explicitly reconstruct a coarse base model and a refined model, here we use the refined model for evaluation.

\noindent\textbf{Seeing from pose angle.} In Table~\ref{tab:eval_wild_a}~\ref{tab:eval_lab_a} and Figure~\ref{fig:benchmark_chart}-(a/b), we can see that most of the methods perform well in frontal views, but degraded for the side views.  PRNet\cite{feng2018joint}, 3DDFA\_v2\cite{guo2020towards}, and MGCNet\cite{shang2020self} are the few methods that are relatively stable for side views. If we only take frontal images (pose angle in $0^\circ-30^\circ$) into account, 3DDFAv2 is the leader, together with PRNet, FaceScape, DECA, and MGCNet are the leading $5$ methods in both the FS-Wild and FS-Lab benchmark while maintaining $>80\%$ complete rates. If we only take side-view images (pose angle in $30^\circ-90^\circ$) into account, MGCNet is the leader, together with 3DDFAv2, PRNet, DECA, and FaceScape are the leading $5$ methods in both the FS-Wild and FS-Lab benchmark while maintaining $>80\%$ complete rates. We can see that all these leading methods are using parametric faces as constraints, and we consider that the data augmentation design in 3DDFAv2 is one of the key reasons for its leading performance. MGCNet achieves the best for side-view faces but falls slightly behind for front-view faces, and we think the reason is that MGCNet is trained with multi-view images and is focusing more on side-view correspondences.

\noindent\textbf{Seeing from focal length.} In Table~\ref{tab:eval_lab_f} and Figure~\ref{fig:benchmark_chart}-(c), we can see that some methods perform better for long focal while others do the opposite.  The reason is that some methods assume the orthogonal projection camera that is more close to the long focal camera, while others assume a perspective projection camera with a pre-defined focal length. For long focal lengths, PRNet, 3DDFAv2, MGCNet, DECA, and FaceScape are 5 leading methods while maintaining $>80\%$ complete rates. The leading 5 methods of short focal lengths are the same as that of long focal lengths, however, the performances of all 5 methods degrade in for the short focal lengths. We consider that SVFR for wide-range focal, especially for short focal lengths, is still a challenging problem.

\subsection{Comparison with Other Benchmarks}

Early benchmarks for 3D face reconstruction use fitted 3D landmarks\cite{bulat2017far} or fitted 3DMM\cite{booth20183d, booth20173d} for evaluation.  Considering the accuracy of fitted 3DMM is low, latter benchmarks choose to capture RGBD data for evaluation\cite{RingNet, feng2018evaluation}.  NoW benchmark\cite{RingNet} and Feng \etal's benchmark\cite{feng2018evaluation} are the only large-scale reconstructed benchmarks for 3D face reconstruction. NoW benchmark contains $2054$ images of $100$ subjects captured with an iPhone X, and a separate 3D head scan for each subject. Feng \etal's benchmark contains $2000$ images of $125$ subjects captured by an RGB camera and a head scanner.  Both benchmarks take various head poses, expressions, and lighting conditions into account, while the accuracy of ground-truth shape is limited to the hand-hold 3D scanner and the aligning algorithm.  Compared to these methods, our benchmark provides extremely well-aligned and fidelity 3D shapes captured by a multi-view camera array (error $< 0.2mm$) as ground truth and provides in-the-wild data and in-the-lab data for a comprehensive evaluation. Our benchmark considers comparing methods according to poses, focal length, and expressions, which reveals more challenges.  NoW benchmark aims at evaluating the accuracy of a neutralized 3D face, so it requires the method to be a parametric method that can be neutralized.  By contrast, our benchmark aims at evaluating the accuracy of a 3D face at the moment the photo was taken, so the face may be in any facial expression and the method does not need to be parametric. Besides, our benchmark and the previous benchmarks are complementary in terms of race: the subjects of our benchmark are mostly oriental faces, while the subjects of NoW and Feng \etal's benchmark are mostly Western faces. 
The comparison between the facescape benchmark and previous benchmarks is summarized in Table~\ref{tab:benchmark_compare}.

\begin{table*}[]
\centering
\caption{Comparison with previous benchmarks.}
\vspace{-0.1in}
\begin{tabular}{@{}lcccc@{}}
\toprule
benchmark             & NoW\cite{RingNet}               & Feng \etal\cite{feng2018evaluation}            & FS-Lab              & FS-Wild                     \\ \midrule
Subject Amount        & 100            & 135                  & 20                  & 400                             \\
Accuracy of GT        & high              & high               & very high           & very high                       \\
Variety of Identity   & western faces     & western faces     & oriental faces      & oriental faces                  \\
Variety of Expression & daily expressions & daily expressions & 20 specified  expressions & daily expressions               \\
Variety of Enviroment & street scenes      & indoor scenes  & laboratory studio         & in-the-wild as CelebA\cite{lee2020maskgan})    \\
Multiple Focal Length & No      & No  & No         & three types    \\
Sense of Reality   & real-captured     & real-captured  & real-captured       & somewhat photo-realistic \\
Neutralized Face   & required     & not required  & not required       & not required \\ \bottomrule
\end{tabular}
\label{tab:benchmark_compare}
\end{table*}

\subsection{Limitations}

\noindent\textbf{Image synthesis artifacts.}
Although we have tried our best to synthesize realistic images, there is still a gap between the synthetic face and the real-captured images. Specifically, in some cases, the lighting model used in the composite image is different from the lighting in the in-the-wild image, and in other cases, the synthesized highlight on the face appears unnatural.

\noindent\textbf{Racial bias.}
As most of the faces in our benchmark are collected from Asians, our benchmark may not be able to accurately assess the accuracy of reconstruction of the appearance of other populations. With more releases of high-quality 3D face data of various races in the future, a comprehensive evaluation without racial bias can be established by merging multiple data sets into our framework.

\section{Conclusion and Future Work}

We present a large-scale detailed 3D facial dataset, FaceScape. Compared to previous public large-scale 3D face datasets, FaceScape provides the highest geometry quality and the largest model amount, and also a comprehensive benchmark for evaluating single-view face reconstruction.
We explore the prediction of a detailed riggable 3D face model from a single image and achieve high fidelity in dynamic detail synthesis.  We believe the release of FaceScape will spur future research on 3D facial modeling and parsing.

There are still some unsolved problems in our work. In terms of data sets, most of our 3D faces are collected from Asians, so our data distribution has racial bias. In the future, with more releases of high-quality 3D face data of various races, racial bias can be eliminated by merging multiple data sets. Our benchmark shares the same problem and solutions about racial distribution bias. Regarding algorithms, our methods explore the potential to model the dynamic geometry details, however, there is still room to improve regarding accuracy and generalization. Specifically, our SVFR method recovers accurate large-scale geometry from landmarks and plausible small-scale geometry from texture-displacement correspondence but fails to recover an accurate middle-scale geometry. Besides, the performance of our SVFR method will degrade when the face is in a side view. We hope that the methods and benchmarks proposed in this paper will inspire further progress in SVFR research.

\ifCLASSOPTIONcaptionsoff
  \newpage
\fi



%



\bibliographystyle{IEEEtran}
\bibliography{bib_face}

%

\begin{IEEEbiography}
	[{\includegraphics[width=1in,height=1.25in,clip,keepaspectratio]{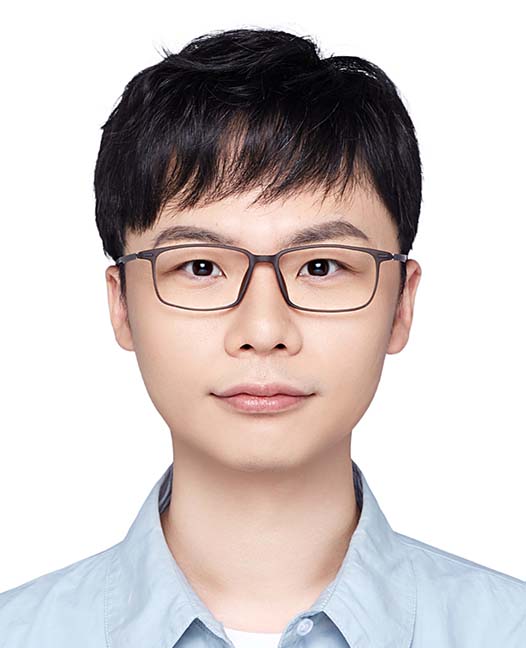}}]
	{Hao Zhu} received the Ph.D. and B.S. degrees from Nanjing University, Nanjing, China. He was a visiting scholar at the University of Kentucky and a visiting student at Tsinghua University.  He is currently an assistant professor at Nanjing University.  His current research interests include computer vision and deep learning, especially 3D reconstruction and 3D vision. 
\end{IEEEbiography}

\vspace{-0.3in}
\begin{IEEEbiography}
	[{\includegraphics[width=1in,height=1.25in,clip,keepaspectratio]{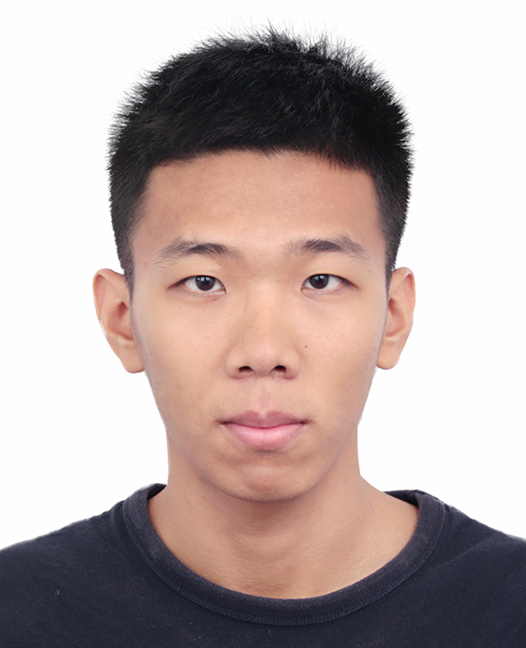}}]
	{Haotian Yang} received the M.S. degree and B.S. degree from the School of Electronic Science and Engineering, Nanjing University, China, in 2021 and 2018 respectively. His research interests include computer vision and computer graphics. 
\end{IEEEbiography}

\vspace{-0.3in}
\begin{IEEEbiography}
	[{\includegraphics[width=1in,height=1.25in,clip,keepaspectratio]{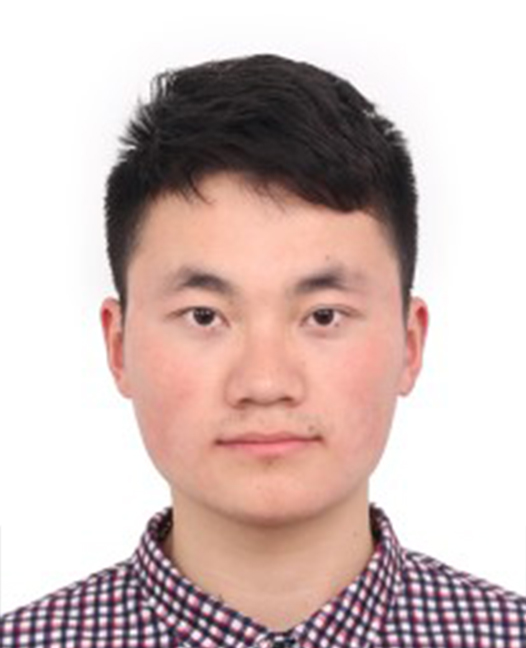}}]
	{Longwei Guo} received a B.S. degree from Nanjing University, Nanjing, China, in 2020, where he is currently pursuing an M.S. degree with the School of Electronic Science and Engineering. His current research interests include computer vision and computer graphics. 
\end{IEEEbiography}

\vspace{-0.3in}
\begin{IEEEbiography}
	[{\includegraphics[width=1in,height=1.25in,clip,keepaspectratio]{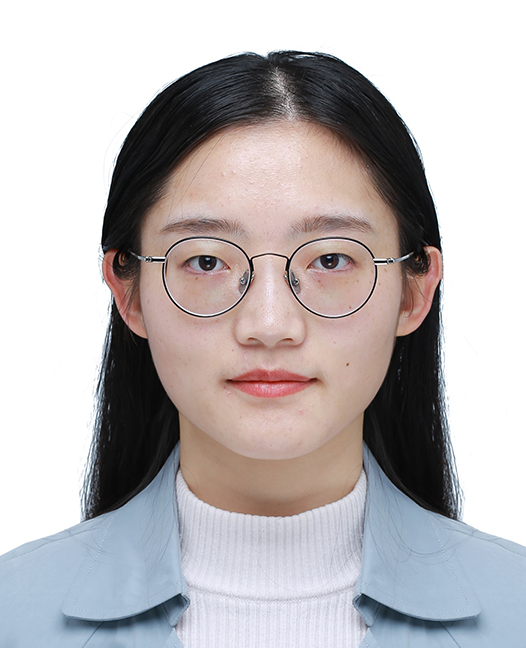}}]
	{Yidi Zhang} received the M.S. degree and B.S. degree from the School of Electronic Science and Engineering, Nanjing University, China, in 2021 and 2018 respectively. His research interests include computer vision and computer graphics. 
\end{IEEEbiography}

\vspace{-0.3in}
\begin{IEEEbiography}
	[{\includegraphics[width=1in,height=1.25in,clip,keepaspectratio]{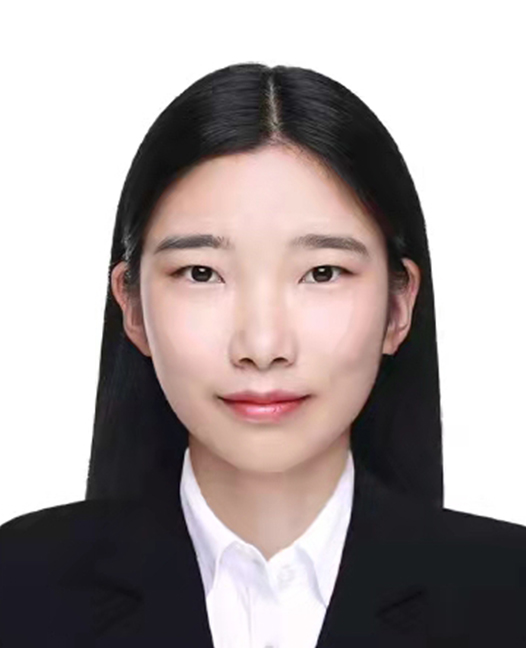}}]
	{Yanru Wang} received the M.S. degree from the School of Electronic Science and Engineering, Nanjing University, China, in 2020. She received a B.S. degree from the Hefei University of Technology in 2017. Her research interests include computer vision and computer graphics. 
\end{IEEEbiography}

\vspace{-0.3in}
\begin{IEEEbiography}
	[{\includegraphics[width=1in,height=1.25in,clip,keepaspectratio]{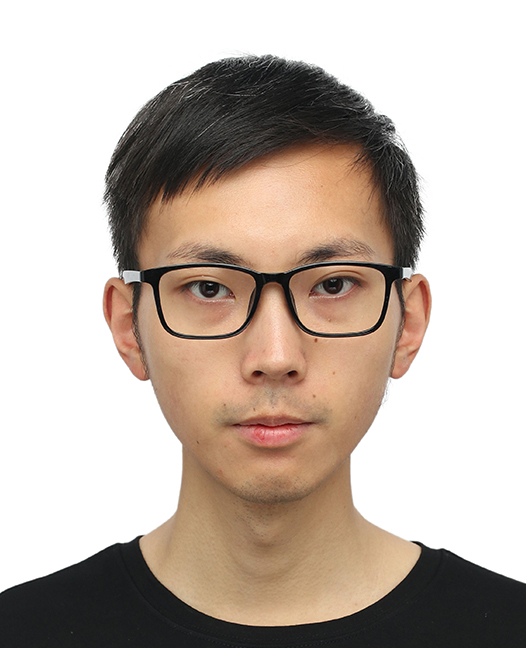}}]
	{Mingkai Huang} received the M.S. degree and B.S. degree from the School of Electronic Science and Engineering, Nanjing University, China, in 2020 and 2017 respectively. His research interests include computer vision and computer graphics. 
\end{IEEEbiography}

\vspace{-0.3in}
\begin{IEEEbiography}
	[{\includegraphics[width=1in,height=1.25in,clip,keepaspectratio]{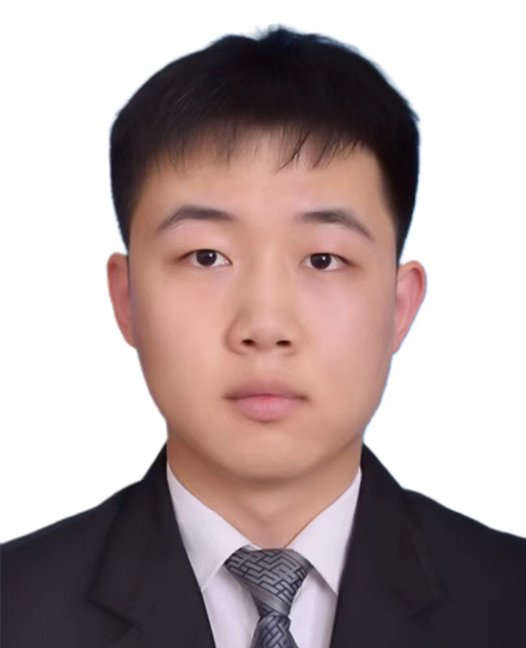}}]
	{Menghua Wu} is a postgraduate student at the School of Electronic Science and Engineering, Nanjing University, China. He received a B.S. degree from Xidian University in 2021. His research interests include computer vision and computer graphics. 
\end{IEEEbiography}

\vspace{-0.3in}
\begin{IEEEbiography}
	[{\includegraphics[width=1in,height=1.25in,clip,keepaspectratio]{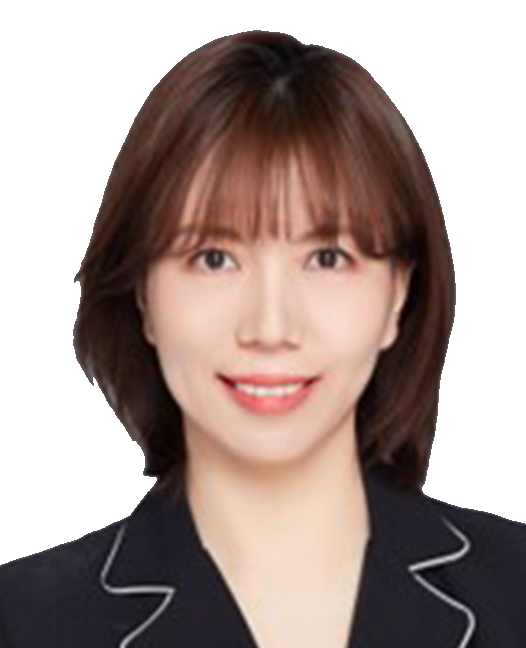}}]
	{Qiu Shen} received the B.S. and Ph.D. degree from the University of Science and Technology of China, in 2004 and 2009 respectively. From 2009 to 2016, he has been with Huawei 2012 Lab and Nanjing University of Aeronautics and Astronautics. She is now on the faculty of Electronic Science and Engineering School, Nanjing University, China. Her current research focuses on next-generation video coding, collaborative video compression and analysis, and vision model for virtual reality.
\end{IEEEbiography}

\vspace{-0.3in}
\begin{IEEEbiography}
	[{\includegraphics[width=1in,height=1.25in,clip,keepaspectratio]{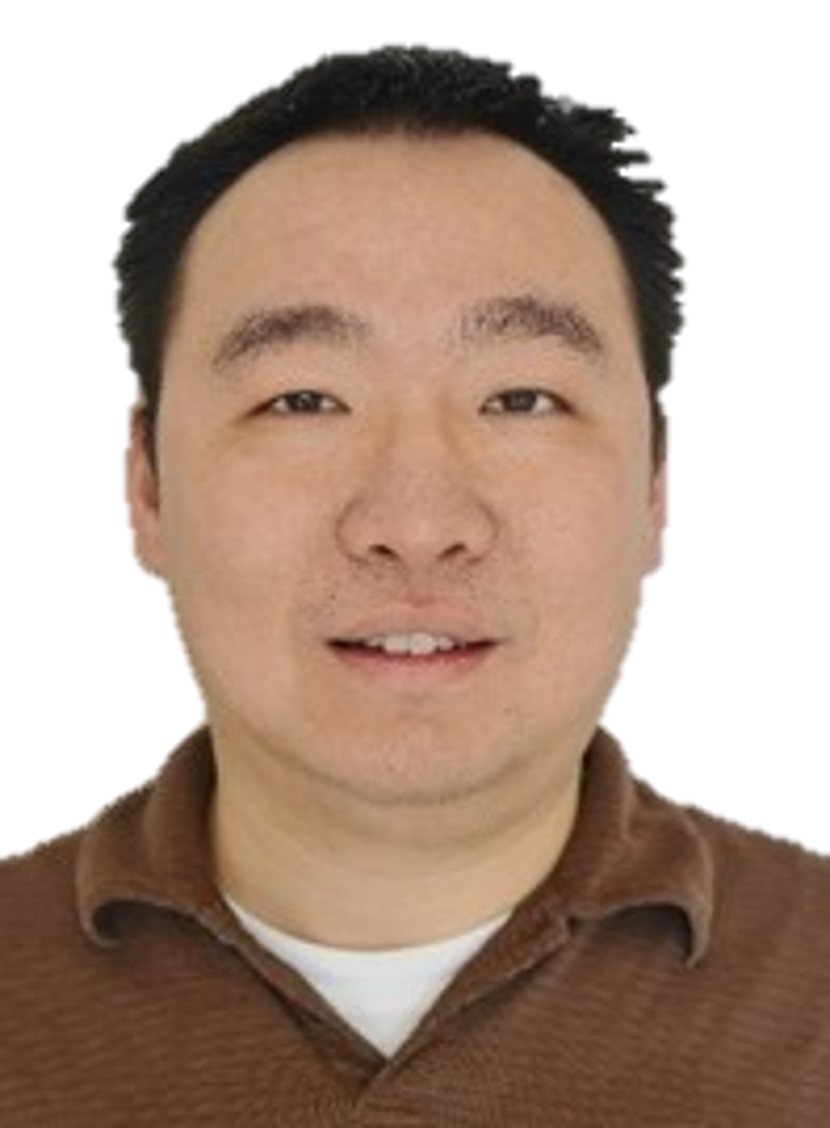}}]{Ruigang Yang}
	is currently a full professor of Computer Science at the University of Kentucky. He obtained his Ph.D. degree from the University of North Carolina at Chapel Hill and his MS degree from Columbia University. His research interests span computer graphics and computer vision, in particular in 3D reconstruction and 3D data analysis. He has published over 100 papers, which, according to Google Scholar, have received 
more than 10000 citations with an h-index of 52 (as of 2019). He has received a number of awards, including the US NSF Career award in 2004 and the Dean’s Research Award at the University of Kentucky in 2013. He is a Fellow of IEEE.
\end{IEEEbiography}

\vspace{-0.3in}
\begin{IEEEbiography}
	[{\includegraphics[width=1in,height=1.25in,clip,keepaspectratio]{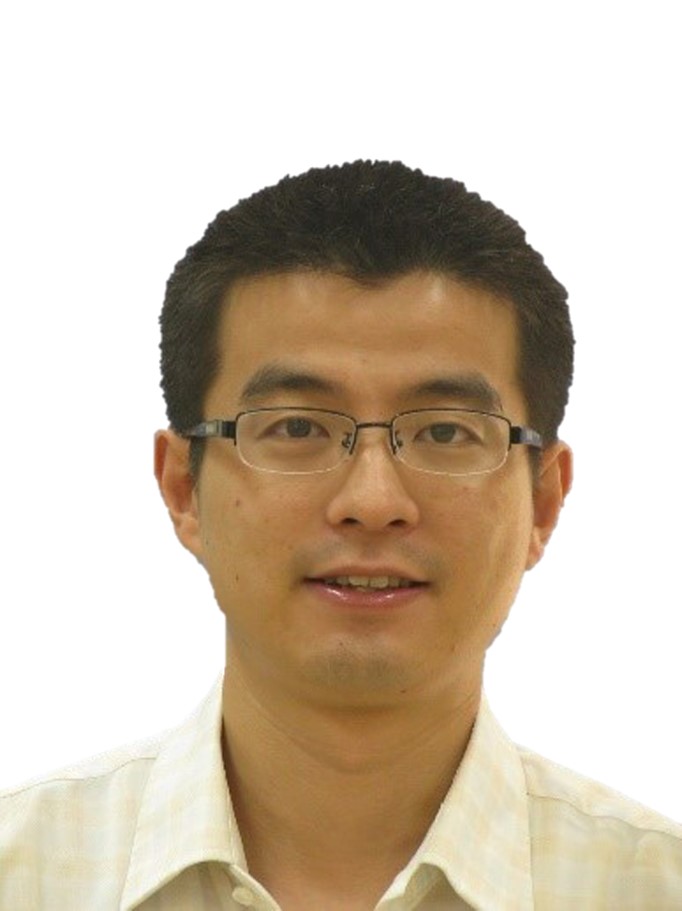}}]
	{Xun Cao} received a B.S. degree from Nanjing University, Nanjing, China, in 2006, and a Ph.D. degree from the Department of Automation, Tsinghua University, Beijing, China, in 2012. He held visiting positions with Philips Research, Aachen, Germany, in 2008, and Microsoft Research Asia, Beijing, from 2009 to 2010. He was a Visiting Scholar with the University of Texas at Austin, Austin, TX, USA, from 2010 to 2011. He is currently a Professor at the School of Electronic Science and Engineering, at Nanjing University. His current research interests include computational photography and image-based modeling and rendering. 
\end{IEEEbiography}








\end{document}